\documentclass[journal]{IEEEtran}

\usepackage{epsfig}
\usepackage{graphicx}
\usepackage{amsmath}
\usepackage{amssymb}
\usepackage{subfigure}
\usepackage{url}
\usepackage{bbm}
\usepackage{algorithm}
\usepackage{algorithmic}
\usepackage{caption}

\newcommand{\tabincell}[2]{\begin{tabular}{@{}#1@{}}#2\end{tabular}}

\ifCLASSINFOpdf
\else
\fi
\begin{document}
%
%
\title{Joint Learning of Siamese CNNs and Temporally Constrained Metrics for Tracklet Association}
%
%

\author{Bing~Wang,~\IEEEmembership{Student Member,~IEEE,}
        Li~Wang,~\IEEEmembership{Member,~IEEE,}
        Bing~Shuai,~\IEEEmembership{Student Member,~IEEE,}
        Zhen~Zuo,~\IEEEmembership{Student Member,~IEEE,}
        Ting~Liu,~\IEEEmembership{Student Member,~IEEE,}
        Kap~Luk~Chan,~\IEEEmembership{Member,~IEEE,}
        and Gang~Wang,~\IEEEmembership{Member,~IEEE}
\thanks{B.~Wang, L.~Wang, B.~Shuai, Z.~Zuo, T~Liu, K.~L.~Chan, and G.~Wang are with the School of Electrical and Electronics Engineering, Nanyang Technological University. 50 Nanyang Avenue, Singapore, 639798 (e-mail: wang0775@e.ntu.edu.sg, wa0002li@e.ntu.edu.sg, bshuai001@e.ntu.edu.sg, zzuo1@e.ntu.edu.sg, liut0016@e.ntu.edu.sg, eklchan@ntu.edu.sg, wanggang@ntu.edu.sg).}}

\maketitle


\begin{abstract}
    In this paper, we study the challenging problem of multi-object tracking in a complex scene captured by a single camera. Different from the existing tracklet association-based tracking methods, we propose a novel and efficient way to obtain discriminative appearance-based tracklet affinity models. Our proposed method jointly learns the convolutional neural networks (CNNs) and temporally constrained metrics. In our method, a Siamese convolutional neural network (CNN) is first pre-trained on the auxiliary data. Then the Siamese CNN and temporally constrained metrics are jointly learned online to construct the appearance-based tracklet affinity models. The proposed method can jointly learn the hierarchical deep features and temporally constrained  segment-wise metrics under a unified framework. For reliable association between tracklets, a novel loss function incorporating temporally constrained multi-task learning mechanism is proposed. By employing the proposed method, tracklet association can be accomplished even in challenging situations. Moreover, a new dataset with 40 fully annotated sequences is created to facilitate the tracking evaluation. Experimental results on five public datasets and the new large-scale dataset show that our method outperforms several state-of-the-art approaches in multi-object tracking.
\end{abstract}


%
\IEEEpeerreviewmaketitle

\section{Introduction}

\IEEEPARstart{M}{ulti-object} tracking in real scenes is an important topic in computer vision, due to its demands in many essential applications such as surveillance, robotics, traffic safety and entertainment. As the seminal achievements were obtained in object detection \cite{Dalal,Wang1,Felzenszwalb2}, tracklet association-based tracking methods \cite{Kuo,Yang2,Dicle,Wen,Wang111} have become popular recently. These methods usually include two key components: 1) A tracklet affinity model that estimates the linking probability between tracklets (track fragments), which is usually based on the combination of multiple cues (motion and appearance cues); 2) A global optimization framework for tracklet association, which is usually formulated as a maximum a posterior problem (MAP).

Even though some state-of-the-art methods \cite{Kuo,Dicle,Bae} have achieved much progress in constructing discriminative appearance and motion based tracklet affinity models, problems such as track fragmentation and identity switch still cannot be well handled, especially under difficult situations where the appearance or motion of an object changes abruptly and significantly. Some of state-of-the-art tracklet association-based multi-object tracking methods \cite{Kuo,Bae} make use of image representations which are not well-suited for constructing robust appearance-based tracklet affinity models. These methods usually utilize pre-selected features, such as HOG features \cite{Dalal}, local binary patterns \cite{Wang1}, or color histograms, which are not ``tailor-made" for the tracked objects in question. Recently, deep convolutional neural network architectures have been successfully applied to many challenging tasks, such as image classification \cite{Krizhevsky} and object detection \cite{Girshick}, and been reported highly promising results. The core to the deep convolutional neural network's success is to take the advantage of deep architectures to learn richer hierarchical features through multiple nonlinear transformations. Hence, we adopt the deep convolutional neural network for multi-object tracking in this work.

\begin{figure*}[t]
\begin{center}
\includegraphics[width=1.0\linewidth]{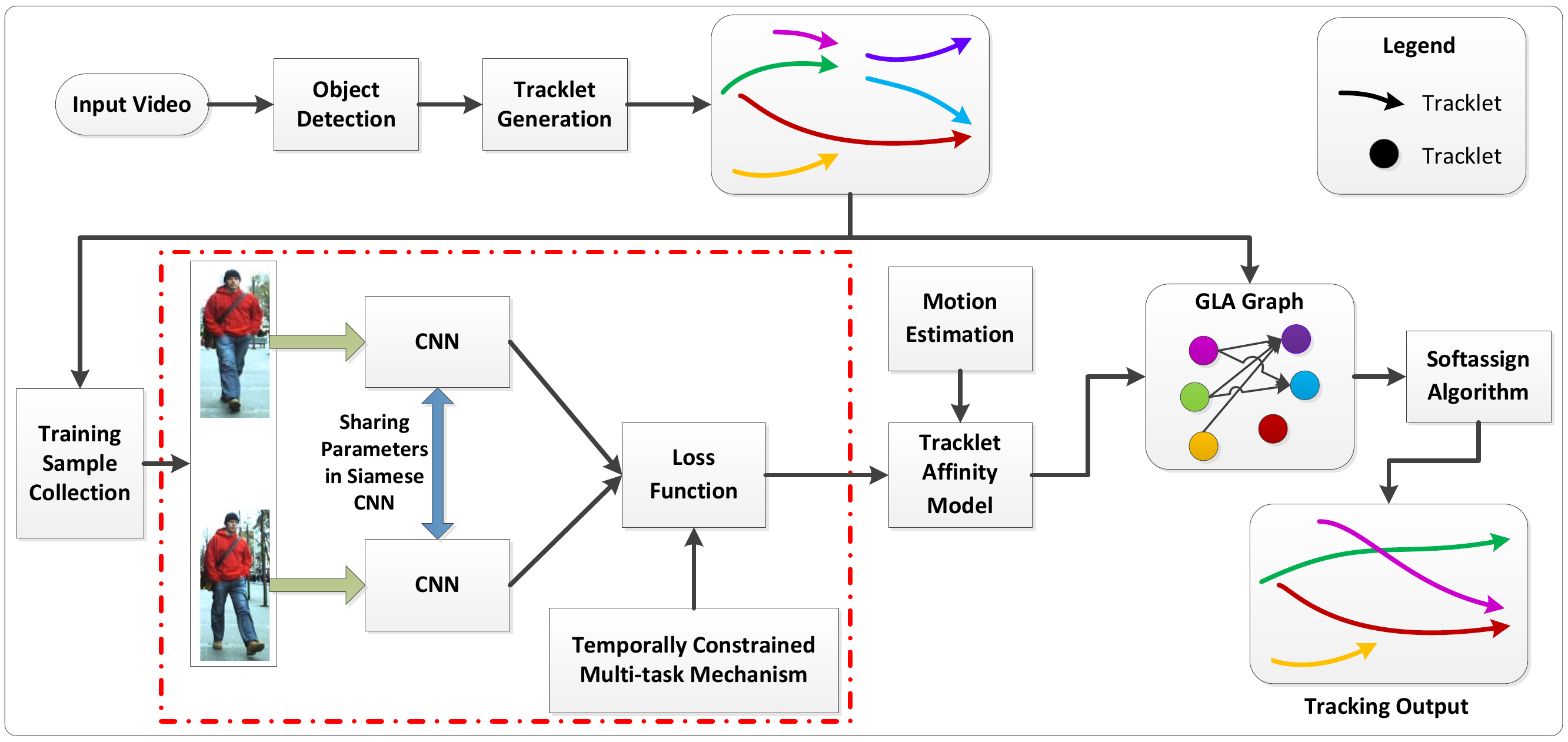}
\end{center}
   \caption{Tracking framework of our method. In the Generalized Linear Assignment (GLA) \cite{Shmoys} graph, each node denotes a reliable tracklet; each edge denotes a possible link between two tracklets. Our method jointly learns the parameters of the Siamese CNN and the temporally constrained metrics for tracklet affinity model, as shown in the red-dashed box, which estimates the linking probability between two tracklets in the GLA graph. The tracking results are obtained by combinatorial optimization using the softassign algorithm \cite{Gold}.}
\label{fig:1}
\end{figure*}

Traditional deep neural networks \cite{Krizhevsky,Oquab} are designed for the classification task. Here, we aim to associate tracklets by joint learning of the convolutional neural networks and the appearance-based tracklet affinity models. This joint optimization will maximize their capacity for solving tracklet association problems. Hence, we propose to jointly learn the Siamese convolutional neural network, which consists of two sub-networks (see Figure \ref{fig:1}), and appearance-based tracklet affinity models, so that the appearance-based affinity models and the ``tailor-made" hierarchical features for tracked targets are learned simultaneously and coherently. Furthermore, based on the analysis of the characteristics of the sequential data stream, a novel temporally constrained multi-task learning mechanism is proposed to be added to the objective function. This makes the deep architectures more effective in tackling the tracklet association problem. Although deep architectures have been employed in single object tracking \cite{Wang2,Li01,Wang3,Zhang03,Hong01}, we explore deep architectures for multi-object tracking in this work.

The proposed framework in this paper is shown in Figure \ref{fig:1}. Given a video input, we first detect objects in each frame by a pre-trained object detector, such as the popular DPM detector \cite{Felzenszwalb2}. Then a dual-threshold strategy \cite{Huang2} is employed to generate reliable tracklets. The Siamese CNN is first pre-trained on the auxiliary data offline. Subsequently, the Siamese CNN and temporally constrained metrics are jointly learned online for tracklet affinity models by using the online collected training samples among the reliable tracklets. Finally, the tracklet association problem is formulated as a Generalized Linear Assignment (GLA) problem, which is solved by the softassign algorithm \cite{Gold}. The final trajectories of multiple objects are obtained after a trajectory recovery process.

The contributions of this paper can be summarized as: (1) We propose a unified deep model for jointly learning ``tailor-made" hierarchical features for currently tracked objects and temporally constrained segment-wise metrics for tracklet affinity models. With this deep model, the feature learning and the discriminative tracklet affinity model learning can efficiently interact with each other, maximizing their performance co-operatively. (2) A novel temporally constrained multi-task learning mechanism is proposed to be embedded into the last layer of the unified deep neural network, which makes it more effective to learn appearance-based affinity model for tracklet association. (3) A new large-scale dataset with 40 diverse fully annotated sequences is built to facilitate performance evaluation. This new dataset includes 24,882 frames and 246,330 annotated bounding boxes of pedestrians.

The remainder of this paper is organized as follows: The unified deep model is introduced in Section \ref{sec:2}. Section \ref{sec:3} describes the tracklet association framework. The new large-scale dataset and experimental results are introduced in Section \ref{sec:4}. Section \ref{sec:5} concludes the paper.

\section{The Unified Deep Model} \label{sec:2}

In this section, we explain how the unified deep model is designed for jointly learning hierarchical features and temporally constrained metrics for tracklet association.

\subsection{The Architecture}

A deep neural network usually works in a standalone mode for most of computer vision tasks, such as image classification, object recognition and detection. The input and output of the deep neural network in this mode are a sample and a predicted label respectively. However, for the tracklet association problem, the objective is to estimate the tracklet affinities between two tracklets to decide whether they belong to the same object. Hence, the ``sample $\rightarrow$ label" mode deep neural network is not applicable to the tracklet association problem. To deal with this problem, we propose to create a Siamese deep neural network, which consists of two sub-networks working in a ``sample pair $\rightarrow$ similarity" mode.

The structure of the Siamese convolution neural network (CNN) is shown in Figure \ref{fig:1} (red-dashed box). Given two target images, they are first warped to a fixed 96 $\times$ 96 patch and presented to the Siamese CNN. The Siamese CNN is composed of two sub convolutional neural networks (CNNs), as shown in Figure \ref{fig:1} (red-dashed box). A novel metric learning based loss function is proposed for learning this Siamese CNN. Moreover, the Siamese CNN has their two sub-CNNs sharing the same parameters, \emph{i.e.}, weights and biases.

\begin{figure}[htb]
\begin{center}
\includegraphics[width=1.0\linewidth]{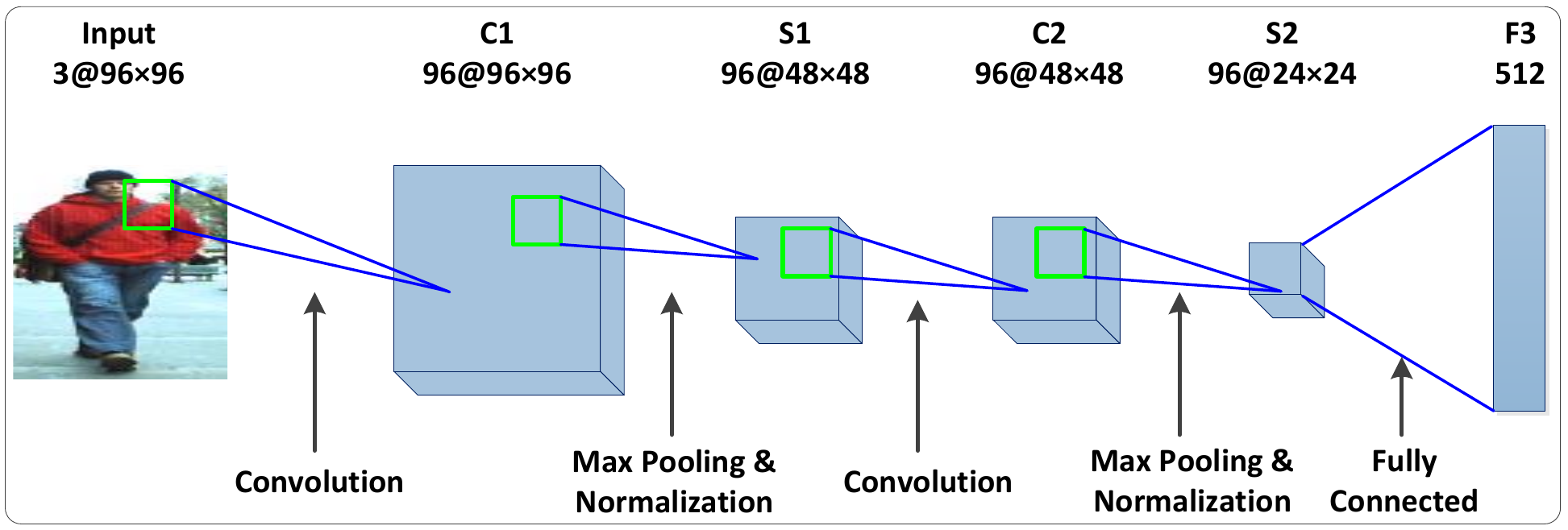}
\end{center}
   \caption{The structure of 5-layer sub-CNN used in the unified deep model.}
\label{fig:3}
\end{figure}

The sub-CNN in the unified deep model consists of 2 convolutional layers (C1 and C2), 2 max pooling layers (S1 and S2) and a fully connected layer (F3), as shown in Figure \ref{fig:3}. The number of channels of convolutional and max pooling layers are both 96. The output of the sub-CNN is a feature vector of 512 dimensions. A cross-channel normalization unit is included in each pooling layer. The convolutional layer output has the same size as the input by zero padding of the input data. The filter sizes of C1 and C2 layers are 7 $\times$ 7 and 5 $\times$ 5 respectively. The activation function for each layer in the CNN is ReLU neuron \cite{Krizhevsky}.

\subsection{Loss Function and Temporally Constrained Metric Learning}

As we can see in Figure \ref{fig:1} (red-dashed box), the Siamese CNN consists of two basic components: two sub-CNNs and a loss function. The loss function converts the difference between the paired input samples into a margin-based loss.

The relative distance between an input sample pair used in the loss function, parameterized as a Mahalanobis distance, is defined as:
\begin{align}
{{\|x_i-x_j\|}_M}^2= {(x_i-x_j)}^T M (x_i-x_j), \quad i\neq j \label{eq:1}
\end{align}
where $x_i$ and $x_j$ are two 512-dimensional feature vectors obtained from the last layer of the two sub-CNNs; and $M$ is a positive semidefinite matrix.

Before introducing the proposed loss function with the temporally constrained multi-task learning mechanism, we first present the loss function with common metric learning. Given training samples, we aim to minimize the following loss:
\begin{align}
& \min_M \frac{\lambda}{2}{{\|M-I\|}_F}^2 + C \sum_{i,j} \max(0,b-l_{i,j}[1-{{\|x_i-x_j\|}_M}^2]) \notag \\
& s.t. \quad M\succeq 0, i\neq j \label{eq:2}
\end{align}
where $\lambda$ is a regularization parameter; ${\|\cdot \|}_F$ denotes the Frobenius norm of a matrix; $C$ is the weight parameter of the empirical loss; $b$ is a constant value satisfying $0\leq b \leq 1$, which represents the decision margin; $l_{i,j}$ is a label that equals to 1 when $x_i$ and $x_j$ are of the same object and -1 otherwise; and $M\succeq 0$ means that $M$ is a positive semidefinite matrix.

Nevertheless, object appearance can vary a lot in the entire video sequence. It is undesirable to use the same metric to estimate the tracklet affinities over the entire video sequence. In this paper, segment-wise metrics are proposed to be learned within each short-time segment, known as local segment. Meanwhile, to capture the common discriminative information shared by all the segments, a multi-task learning mechanism is proposed to be embedded into the loss function for learning the segment-wise and common metrics simultaneously. Moreover, segments in videos are temporal sequences. Temporally close segments should share more information. Hence, we propose a multi-task learning method incorporating temporal constraints for this learning problem:
\begin{align}
\min_{M_0,...,M_n} & \bigg( \frac{\lambda_0}{2}{{\|M_0-I\|}_F}^2 + \sum_{t=2}^n \frac{\eta}{2} {{\|M_t-M_{t-1}\|}_F}^2  + \notag \\
& \sum_{t=1}^n [\frac{\lambda}{2} {{\|M_t\|}_F}^2 + C\sum_{i,j} h(x_i,x_j)] \bigg) \notag \\
& s.t. \quad M_0,M_1,...,M_n \succeq 0, i\neq j \label{eq:3}
\end{align}
where $\lambda_0$ and $\lambda$ are the regularization parameters of $M_t$ for $t=0,1,...,n$; $n$ is the total number of segments; $M_0$ is the common metric shared by all the segments; $M_t$ is the segment-wise metric; ${\|\cdot \|}_F$ denotes the Frobenius norm of a matrix; the second term of this loss function is the temporal constraint term, in which $\eta$ is a regularization parameter; $h(x_i,x_j)$ is the empirical loss function; and $C$ is the weight parameter of the empirical loss.

The empirical loss function $h(x_i,x_j)$ used in Equation \eqref{eq:3} is expressed as:
\begin{align}
& h(x_i,x_j)=\max(0,b-l_{i,j}[1-{{\|x_i-x_j\|}_{M_{tot}}}^2]); \label{eq:4} \\
& M_{tot}=M_0+M_t, \quad i\neq j, \notag \\
& {{\|x_i-x_j\|}_{M_{tot}}}^2={(x_i-x_j)}^T (M_0+M_t) (x_i-x_j) \notag
\end{align}
where $b$ is a constant value, which represents the decision margin; $l_{i,j}$ is a label that equals to 1 when $x_i$ and $x_j$ are of the same object and -1 otherwise; $x_i$ and $x_j$ are two 512-dimensional feature vectors obtained from the last layer of the two sub-CNNs; and $M_{tot}$ is the metric used for estimating the relative distance between a sample pair.

\begin{figure}[htb]
\begin{center}
\includegraphics[width=1.0\linewidth]{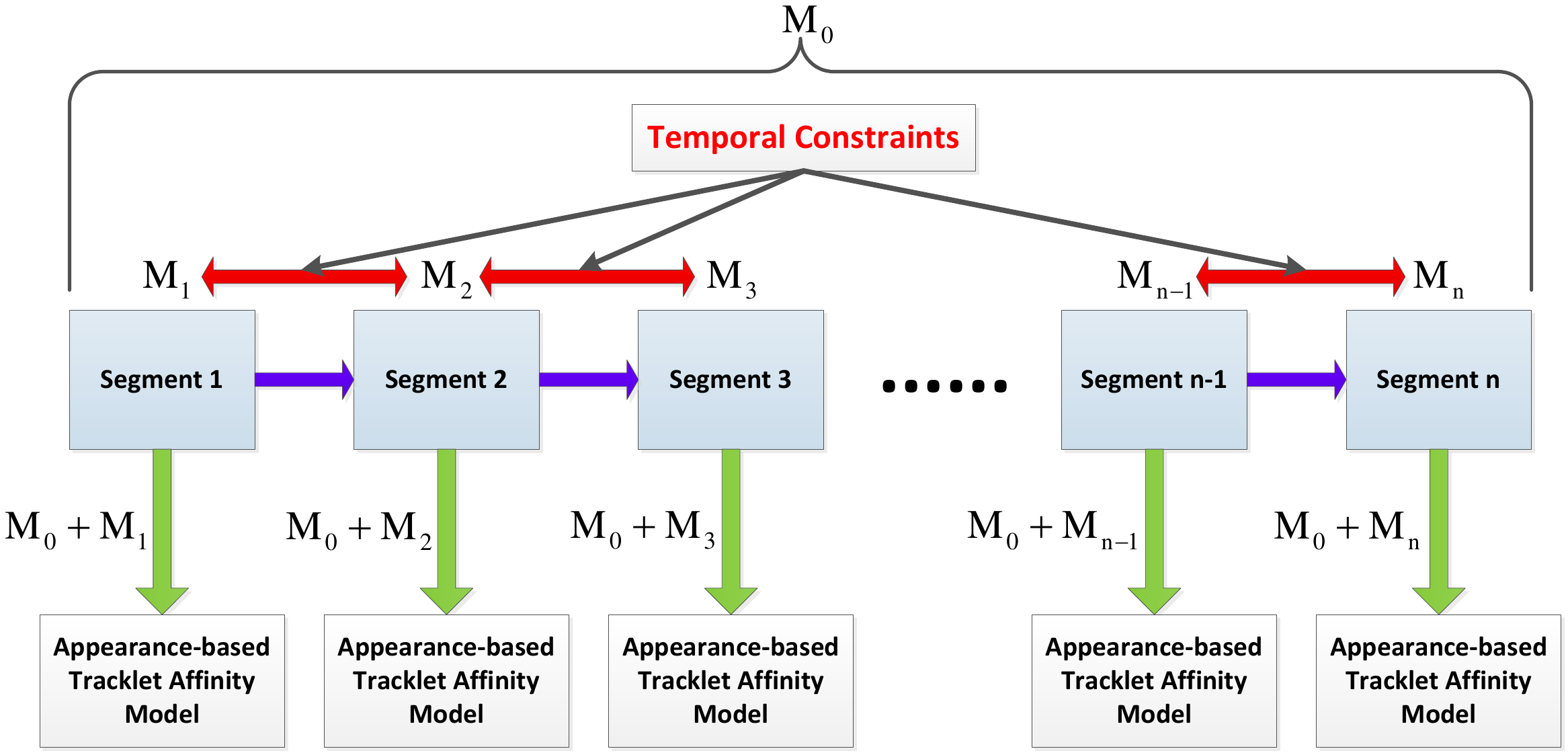}
\end{center}
   \caption{An illustration of the temporally constrained multi-task learning mechanism. $n$ is the total number of the segments and the segments are shown in the temporal space.}
\label{fig:4}
\end{figure}

Intuitively, the common metric $M_0$ represents the shared discriminative information across the entire video sequence and the segment-wise metric $M_{t>0}$ adapt the metric for each local segment. In the proposed objective function, in Equation \eqref{eq:3}, the second term is the temporal constraint term, which accounts for the observation that the neighboring segments sharing more information than the non-neighboring segments (see Figure \ref{fig:4} for an illustration). In the implementation, we use the previous segment-wise metric $M_{t-1}$ in temporal space to initialize the current segment-wise metric $M_t$, due to the assumption that the neighboring segment-wise metrics are more correlated than the non-neighboring ones.

To learn the parameters of the unified deep model, back-propagation (BP) \cite{LeCun} is utilized. The forward propagation function to calculate the loss of the training pairs is presented in Equation \eqref{eq:3}. By differentiating the loss function with respect to the two input samples, we have the gradients. The total gradient for back-propagation is the sum of the contributions from the two samples, which is as follows:
\begin{align}
\nabla G_{total}=2C l_{i,j} (M_{tot}+{M_{tot}}^T)(x_i-x_j) \left( \mathbbm{I} \{g(x_i,x_j)>0\} \right) \label{eq:5}
\end{align}
where
\begin{align}
& M_{tot}=M_0+M_t, \label{eq:6} \\
& g(x_i,x_j)=b-l_{i,j}[1-{{\|x_i-x_j\|}_{M_{tot}}}^2], \label{eq:7} \\
& {{\|x_i-x_j\|}_{M_{tot}}}^2={(x_i-x_j)}^T (M_0+M_t) (x_i-x_j) \label{eq:8}
\end{align}
and $\mathbbm{I}\{\cdot\}$ is the indicator function.

Based on Equations \eqref{eq:3} and \eqref{eq:5}, we can learn the parameters of the unified deep model by stochastic gradient descent \cite{Krizhevsky} via back-propagation. Moreover, the temporally constrained metrics for tracklet affinity models are obtained simultaneously by batch mode stochastic gradient descent.
By differentiating the loss function, as shown in Equation \eqref{eq:3}, with respect to the common metric $M_0$ and the segment-wise metric $M_{t>0}$, the gradients are:
\begin{align}
\nabla G_0 & =\frac{\partial L}{\partial M_0} \notag \\
& =\lambda_0 (M_0-I) + C \sum_{i,j} l_{i,j} A_{i,j} \left( \mathbbm{I} \{g(x_i,x_j)>0\} \right); \label{eq:9} \\
\nabla G_t & =\frac{\partial L}{\partial M_t} \notag \\
& = \left\{
 \begin{array}{ll}
             \lambda M_t + C \sum_{i,j} l_{i,j} A_{i,j} \left( \mathbbm{I} \{g(x_i,x_j)>0\} \right), \\
             \quad \quad \quad \quad \quad \quad \quad \quad \quad \quad \text{if} \ \ t=1; \\ \\
             \Big( \eta (M_t-M_{t-1})+\lambda M_t + \\
             C \sum_{i,j} l_{i,j} A_{i,j} \left( \mathbbm{I} \{g(x_i,x_j)>0\} \right) \Big), \\
             \quad \quad \quad \quad \quad \quad \quad \quad \quad \quad \text{otherwise} \ (t>1).
             \end{array}
\right. \label{eq:10}
\end{align}
where
\begin{align}
& A_{i,j}=(x_i-x_j)(x_i-x_j)^T, \label{eq:11} \\
& g(x_i,x_j)=b-l_{i,j}[1-{{\|x_i-x_j\|}_{M_{tot}}}^2], \label{eq:12} \\
& {{\|x_i-x_j\|}_{M_{tot}}}^2={(x_i-x_j)}^T (M_0+M_t) (x_i-x_j) \label{eq:13}
\end{align}

Online training sample collection is an important issue in the learning of the unified deep model. We take the assumptions similar to those as in \cite{Kuo}: (1) detection responses in one tracklet are from the same object; (2) any detection responses in two different tracklets which have overlaps over time are from different objects. The first one is based on the assumption that the tracklets generated by the dual-threshold strategy are reliable; the second one is based on the fact that one target cannot appear at two or more different locations at the same time, known as spatio-temporal conflict. For each tracklet, $\kappa$ strongest detection responses are selected as training samples ($\kappa=4$ in our implementation). Then we use two arbitrarily selected different detection responses from the $\kappa$ strongest responses of $T_i$ as positive training samples, and two detection responses from the $\kappa$ strongest responses of two spatio-temporal conflicted tracklets as negative training samples.

Finally, the common metric $M_0$ and the segment-wise metrics $M_{t>0}$ are obtained simultaneously through a gradient descent rule. The online learning algorithm is summarized in Algorithm \ref{alg:01}.
\begin{align}
M_0= & M_0-\beta \frac{\partial L}{\partial M_0} \label{eq:14} \\
M_t= & M_t-\beta \frac{\partial L}{\partial M_t} \label{eq:15}
\end{align}
where $\beta$ is the learning rate.

Meanwhile, the Siamese CNN is online fine-tuned through back propagating the gradients calculated by Equation \eqref{eq:5}.

\begin{algorithm} \small
\caption{Online Learning Algorithm for Temporally Constrained Metric Learning}
\label{alg:01}
\begin{algorithmic}[1]
\REQUIRE ~~ \\
Feature vectors of online collected training samples $\{{x_i}^t\}$; $i=1,...,n_t$, $n_t$ is the number of the samples within segment $t$; $t=1,...,n$, $n$ is the total number of the segments; and learning rate $\beta$. \\
\ENSURE ~~ \\
The learned metrics: $M_0, M_1,..., M_n$.
\STATE Initialize $M_0=I$ (identity matrix).
\FOR {$t=1,...,n$}
\IF {$t==1$}
\STATE Initialize $M_t=0$.
\ELSE
\STATE Initialize $M_t=M_{t-1}$.
\ENDIF
\STATE Randomly generate the training pairs $\{x_i,x_j,l_{i,j}\}$ from $\{{x_i}^t\}$. $l_{i,j}=1$, if $x_i$ and $x_j$ are from one tracklet; $l_{i,j}=-1$, if $x_i$ and $x_j$ are from two different tracklets which have overlaps over time. A total of $2m$ training pairs in a random order are generated, which includes $m$ positive and $m$ negative pairs.
\FOR {$p=1,...,2m$}
\IF {$l_{i,j}[1-{(x_i-x_j)}^T (M_0+M_t) (x_i-x_j)]>b$}
\STATE $M_0$=$M_0$; $M_t$=$M_t$.
\ELSIF {$l_{i,j}<0$}
\STATE Compute $M_0$ and $M_t$ by Equations \eqref{eq:14} and \eqref{eq:15}.
\ELSE
\STATE $M_0=\pi_{S+}(M_0-\beta \frac{\partial L}{\partial M_0})$;
\STATE $M_t=\pi_{S+}(M_t-\beta \frac{\partial L}{\partial M_t})$; \\ where $\pi_{S+}(A)$ projects matrix $A$ into the positive semidefinite cone.
\ENDIF
\ENDFOR
\ENDFOR
\end{algorithmic}
\end{algorithm}


\section{Tracklet Association Framework} \label{sec:3}

In this section, we present the tracklet association framework, in which, we incorporate the temporally constrained metrics learned by the unified deep model to obtain robust appearance-based tracklet affinity models.

\subsection{Tracklet Association with Generalized Linear Assignment}

To avoid learning tracklet starting and termination probabilities, we formulate the tracklet association problem as a Generalized Linear Assignment (GLA) \cite{Shmoys}, which does not need the source and sink nodes as in conventional network flow optimization \cite{Zhang,Pirsiavash,Butt,Wang111}. Given $N$ tracklets $\{T_1,...,T_N\}$, the Generalized Linear Assignment (GLA) problem is formulated as:
\begin{align}
& \max_X \sum_{i=1}^N \sum_{j=1}^N P(T_i,T_j) X_{ij} \label{eq:16} \\
& s.t. \quad \sum_{i=1}^N X_{ij}\leq 1; \sum_{j=1}^N X_{ij}\leq 1; X_{ij}\in\{0,1\} \notag
\end{align}
where $P(T_i,T_j)$ is the linking probability between $T_i$ and $T_j$. The variable $X_{ij}$ denotes that $T_i$ is the predecessor of $T_j$ in temporal domain when $X_{ij}=1$ and that, they may be merged during the optimization.

\subsection{Tracklet Affinity Measurement}

To solve the Generalized Linear Assignment (GLA) problem in Equation \eqref{eq:16}, we need to estimate the tracklet affinity score, or equivalently, the linking probability, $P(T_i,T_j)$, between two tracklets. The linking probability $P(T_i,T_j)$ is defined based on two cues: motion and appearance.
\begin{align}
P(T_i,T_j)=P_m(T_i,T_j)P_a(T_i,T_j) \label{eq:17}
\end{align}

The motion-based tracklet affinity model $P_m(T_i,T_j)$ is defined as:
\begin{align}
P_m(T_i,T_j)= & \mathcal{N}(p_i^{tail}+v_i^F \Delta t; p_j^{head},\Sigma) \cdot \notag \\
& \mathcal{N}(p_j^{head}+v_j^B \Delta t; p_i^{tail},\Sigma) \label{eq:18}
\end{align}
where $p_i^{tail}$ is the position of the tail response in $T_i$; $p_j^{head}$ is the position of the head response in $T_j$; $v_i^F$ is the forward velocity of $T_i$; $v_j^B$ is the backward velocity of $T_j$; and $\Delta t$ is the time gap between the tail response of $T_i$ and the head response of $T_j$.

In Equation \eqref{eq:18}, the forward velocity $v_i^F$ is estimated from the head to the tail of $T_i$, while the backward velocity $v_j^B$ is estimated from the tail to the head of $T_j$. It is assumed that the difference of the predicted position and the refined position follows a Gaussian distribution.

To estimate the appearance-based tracklet affinity scores, we need to construct the probe set, consisting of the strongest detection response in each tracklet. The probe set is defined as $G=\{g_i\}$, $i=1,...,N_s$, in which $N_s$ is the number of tracklets in a local segment. Each $T_i$ has only one selected $g_i$ in $G$ to represent itself.

The appearance-based tracklet affinity model $P_a(T_i,T_j)$ is defined based on the learned temporally constrained metrics:
\begin{align}
& d_{ij}^{k} = (x_i^{k}-g_j)^T(M_0+M_t)(x_i^{k}-g_j); \notag \\
& d_{ji}^{k'}= (x_j^{k'}-g_i)^T(M_0+M_t)(x_j^{k'}-g_i); \notag \\
& norm_i^k= \sqrt{(\sum_{j=1}^{N_s} d_{ij}^{k})}; \ norm_j^{k'}= \sqrt{(\sum_{i=1}^{N_s} d_{ji}^{k'})}; \notag \\
& d_{ij}= \Big[\sum_k (\frac{d_{ij}^k}{norm_i^k})\Big]/m_i; \ d_{ji}= \Big[\sum_{k'} \frac{d_{ji}^{k'}}{norm_j^{k'}} \Big]/m_j; \notag \\
& P_a(T_i,T_j)=(d_{ij} d_{ji})^{-1} \label{eq:19}
\end{align}
where $x_i^{k}$ denotes the feature vector of the $k$th detection response in $T_i$; $x_j^{k'}$ denotes the feature vector of the $k'$th detection response in $T_j$; $g_i,g_j\in G$; $m_i$ and $m_j$ are the numbers of detection responses of $T_i$ and $T_j$ respectively.

Through Equation \eqref{eq:17}, we can obtain the predecessor-successor matrix $P$ for the objective function \eqref{eq:16}. To achieve fast and accurate convergence, $P$ is normalized by column and a threshold $\omega$ is introduced to ensure that a reliable tracklet association pair has a high affinity score.
\begin{align}
P(T_i,T_j) = \left\{
 \begin{array}{lll}
             P_m(T_i,T_j)P_a(T_i,T_j), \\
             \quad \ if \ P_m(T_i,T_j)P_a(T_i,T_j)\geq \omega \\ \\
             0, \ \text{otherwise} \\
             \end{array}
\right. \label{eq:20}
\end{align}

The Generalized Linear Assignment problem in Equation \eqref{eq:16} can be solved by the softassign algorithm \cite{Gold}. Due to missed detections, there may exist some gaps between adjacent tracklets in each trajectory after tracklet association. Therefore, the final tracking results are obtained through a trajectory interpolation process over gaps based on a linear motion model.


\section{Experiments} \label{sec:4}
\subsection{Datasets}

\subsubsection{Public Datasets}
To evaluate the multi-object tracking performance of the proposed method, experiments are conducted on five publicly available datasets: PETS 2009 \cite{Ferryman}, Town Centre \cite{Benfold}, Parking Lot \cite{Shu}, ETH Mobile scene \cite{Ess} and MOTChallenge \cite{Leal-Taixe03}.

\subsubsection{New Large-Scale Dataset}

As generally known, a representative dataset is a key component in comprehensive performance evaluation. In recent years, several public datasets have been published for multi-target tracking, such as TUD \cite{Andriluka2}, Town Centre\cite{Benfold}, ParkingLot\cite{Shu}, ETH\cite{Ess} and PETS\cite{Ferryman}. Nevertheless, for most of the public datasets, the number of testing sequences is limited. The sequences usually lack sufficient diversity, which makes the tracking evaluation short of desired comprehensiveness. Recently, the KITTI benchmark \cite{Geiger01} was developed via autonomous driving platform for computer vision challenges, which includes stereo, optical flow, visual odometry, object detection and tracking. However, in this benchmark, the camera is only put on the top of a car for all sequences, resulting in less diversity. Moreover, an up-to-date MOTChallenge benchmark \cite{Leal-Taixe03} was very recently made available for multi-target tracking. Compared with other datasets, this MOTChallenge \cite{Leal-Taixe03} is a more comprehensive dataset for multi-target tracking. However, there are still some limitations for this MOTChallenge benchmark. First, 18 testing sequences are included, which are still insufficient to allow for a comprehensive evaluation for multi-target tracking. In the single object tracking area, researchers usually evaluate on more than 50 sequences \cite{Wuyi01} nowadays. Second, the testing sequences of MOTChallenge lack some specific scenarios such as pedestrians with uniforms. Furthermore, the pedestrian view, which is an important factor affecting the appearance models in multi-target tracking, is not analyzed in the MOTChallenge benchmark. Therefore, it is essential to create a more comprehensive dataset, which includes larger collection of sequences and covers more variations of scenarios, for multi-target tracking.

In this work, a new large-scale dataset containing 40 diverse fully annotated sequences is created to facilitate the evaluation. For this dataset, 10 sequences, which contain 5,080 fames and 52,833 annotated boxes of pedestrians, are used for training; 30 sequences, which contain 19,802 frames and 193,497 annotated boxes of pedestrians, are used for testing.

\begin{figure}[!ht]
\centering
\subfigure[India railway]{
\label{fig:subfig_03:a} 
\includegraphics[width=4.25cm,height=3.35cm]{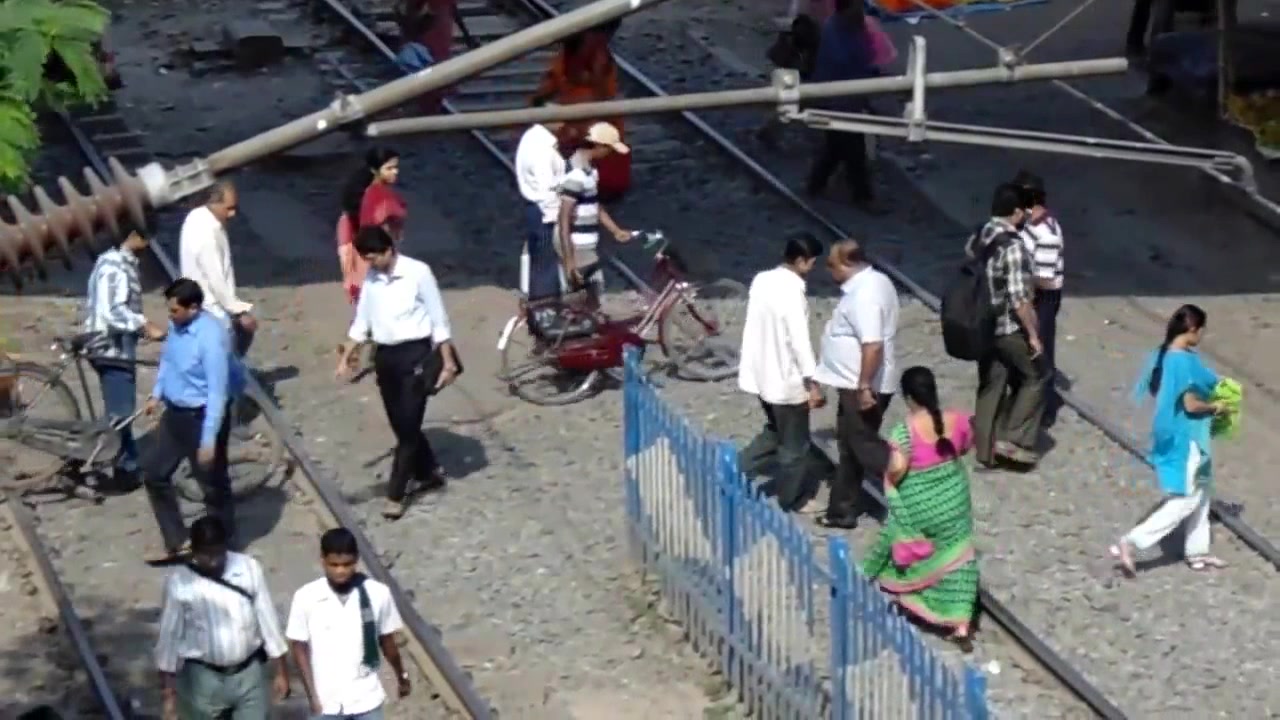}}
\subfigure[Oxford street]{
\label{fig:subfig_03:b} 
\includegraphics[width=4.25cm,height=3.35cm]{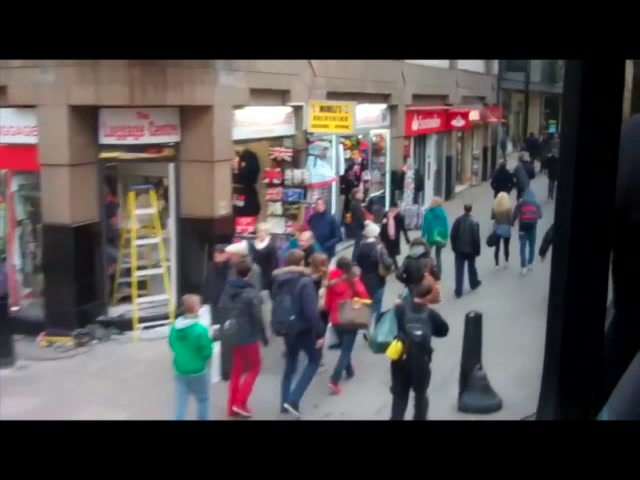}}
\subfigure[College graduation]{
\label{fig:subfig_03:c} 
\includegraphics[width=4.25cm,height=3.35cm]{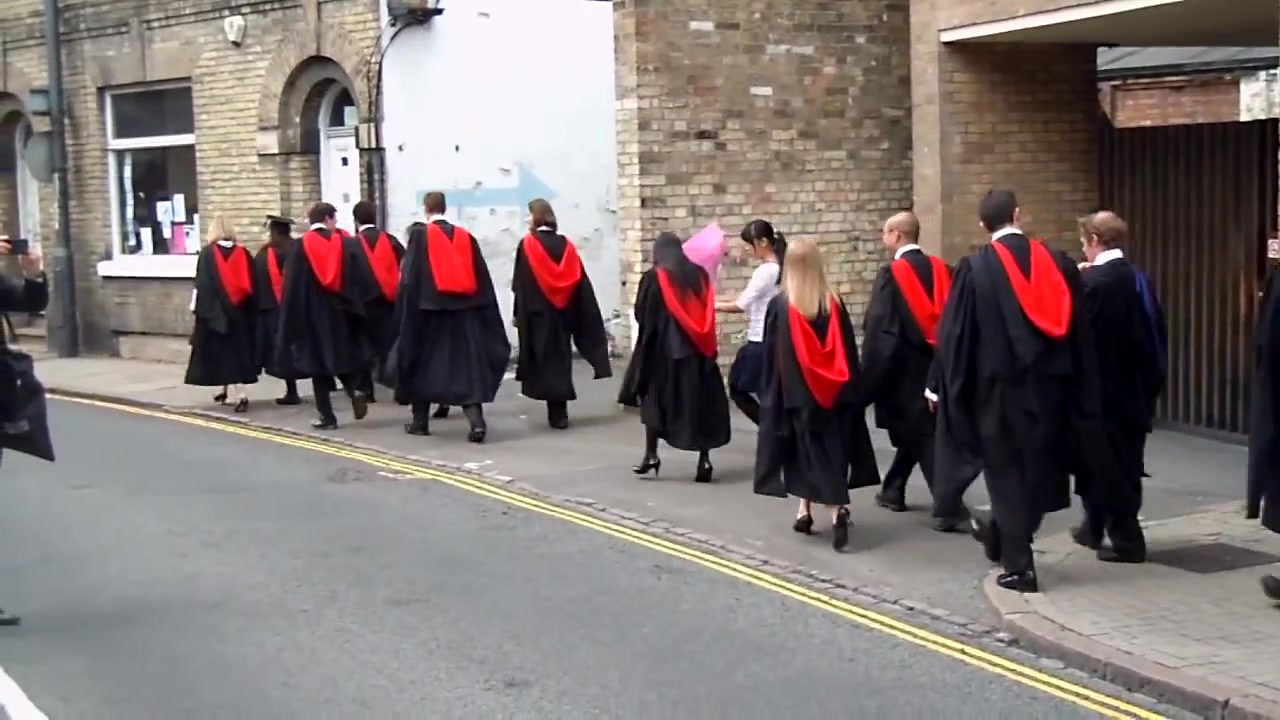}}
\subfigure[Mexico night]{
\label{fig:subfig_03:d} 
\includegraphics[width=4.25cm,height=3.35cm]{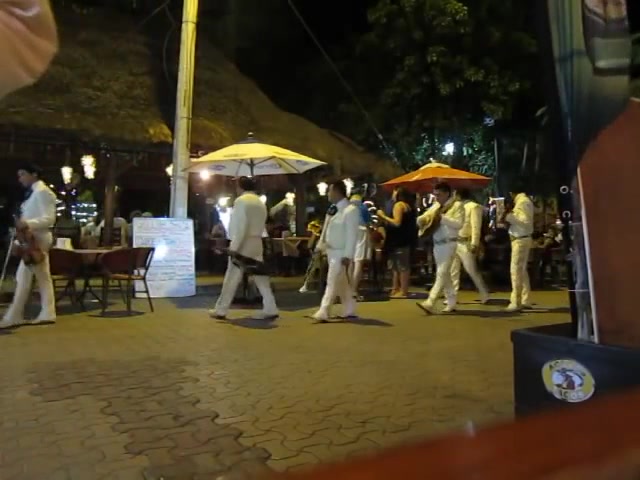}}
\caption{Examples of the new large-scale dataset.}
\label{fig:subfig_03} 
\end{figure}

The new 40 sequences in this new large-scale dataset varies significantly from each other, which can be classified according to the following criteria: (1) \emph{Camera motion.} The camera can be static like most surveillance cameras, or can be moving as held by a person or placed on a moving platform such as a stroller or a car. (2) \emph{Camera viewpoint.} The scenes can be captured from a low position, a medium position (at pedestrian's height), or a high position. (3) \emph{Weather conditions.} Following the settings of MOTChallenge \cite{Leal-Taixe03}, the weather conditions are taken into consideration in this benchmark, which are sunny, cloudy and night. (4) \emph{Pedestrian view.} The dominant pedestrian views in a sequence can be front view, side view or back view. For instance, if a sequence is taken in front of a group of pedestrians, moving towards the camera, the dominant pedestrian view of this sequence is front view. Moreover, the dominant pedestrian views in a sequence can have one, two or all of the views. If one sequence has more pedestrian views, it will be assigned into all the relevant categories. Table \ref{datatb:1} and \ref{datatb:2} show the overview of the training and testing sequences of the new dataset respectively. Figure \ref{fig:subfig_03} shows some examples of this new dataset. As shown in Figure \ref{fig:subfig_03}, this new dataset contains the scenarios that pedestrians with uniforms.

\subsection{Experimental Settings}

\begin{table*}\small
\resizebox{\textwidth}{!}{
\begin{tabular}{|l|c|c|c|c|c|c|c|c|c|c|c|c|}
\hline
Name      & FPS & Resolution & Length & \tabincell{c}{Tracks \\ (in total)} & \tabincell{c}{Boxes of \\ pedestrians} & Density & Camera & Viewpoint & Weather & \tabincell{c}{Front \\ view} &  \tabincell{c}{Side \\ view} & \tabincell{c}{Back \\ view} \\
\hline
tokyo cross   & 25 & 640x480    & 700    & 24     & 6099  & 8.7     & static & low       & cloudy  & yes        & no        & yes             \\
paris cross   & 24 & 1920x1080  & 401    & 34     & 7319  & 18.3    & static & low       & cloudy  & no         & yes       & no              \\
london street & 24 & 1920x1080  & 396    & 23     & 3758  & 9.5     & static & low       & cloudy  & yes        & no        & yes              \\
faneuil hall  & 25 & 1280x720   & 631    & 27     & 3928  & 6.2     & moving & low       & cloudy  & yes        & no        & yes              \\
central park  & 25 & 1280x720   & 600    & 39     & 4400  & 7.3     & static & low       & sunny   & no         & yes       & no               \\
long beach    & 24 & 1280x720   & 402    & 27     & 3599  & 9       & moving & low       & sunny   & yes        & no        & no               \\
oxford street & 25 & 640x480    & 400    & 85     & 5807  & 14.5    & moving & high      & cloudy  & yes        & no        & yes              \\
sunset cross  & 24 & 1280x720   & 300    & 20     & 2683  & 8.9     & static & high      & sunny   & yes        & no        & yes              \\
rain street   & 25 & 640x480    & 800    & 30     & 10062 & 12.6    & moving & low       & cloudy  & yes        & yes       & yes              \\
USC campus    & 30 & 1280x720   & 450    & 37     & 5178  & 11.5    & moving & low       & sunny   & yes        & yes       & yes              \\
\hline
\multicolumn{3}{|c|}{Total training}  & 5080  & 346   & 52833   &        &         &         &        &        &         &                  \\
\hline
\end{tabular}}
\caption{Overview of the training sequences included in the new dataset.}
\label{datatb:1}
\end{table*}

\begin{table*}\small
\resizebox{\textwidth}{!}{
\begin{tabular}{|l|c|c|c|c|c|c|c|c|c|c|c|c|}
\hline
Name & FPS & Resolution & Length & \tabincell{c}{Tracks \\ (in total)} & \tabincell{c}{Boxes of \\ pedestrians} & Density & Camera & Viewpoint & Weather & \tabincell{c}{Front \\ view} &  \tabincell{c}{Side \\ view} & \tabincell{c}{Back \\ view}  \\
\hline
book store             & 24 & 1280x720   & 905    & 78     & 8905   & 9.8     & static & low       & cloudy  & no         & yes       & no                   \\
bronx road             & 24 & 1280x720   & 854    & 64     & 6623   & 7.8     & moving & medium    & sunny   & yes        & no        & yes                  \\
child cross            & 30 & 1280x720   & 518    & 18     & 5389   & 10.4    & static & medium    & cloudy  & yes        & no        & no                   \\
college graduation     & 30 & 1280x720   & 653    & 23     & 7410   & 11.3    & moving & medium    & cloudy  & no         & yes       & yes                  \\
cuba street            & 25 & 1280x720   & 700    & 66     & 7518   & 10.7    & moving & low       & cloudy  & yes        & no        & yes                  \\
footbridge             & 30 & 1280x720   & 678    & 21     & 3195   & 4.7     & moving & low       & sunny   & yes        & no        & yes                  \\
india railway          & 50 & 1280x720   & 450    & 70     & 10799  & 24      & static & high      & sunny   & yes        & yes       & yes                  \\
kuala lumpur airport   & 50 & 1280x720   & 1245   & 30     & 7732   & 6.2     & moving & low       & cloudy  & yes        & no        & no                   \\
latin night            & 25 & 1280x720   & 700    & 23     & 5168   & 7.4     & moving & low       & night   & yes        & no        & yes                  \\
london cross           & 30 & 640x480    & 740    & 30     & 4548   & 6.1     & static & low       & cloudy  & yes        & no        & yes                  \\
london subway          & 30 & 1920x1080  & 361    & 31     & 2424   & 6.7     & static & medium    & cloudy  & yes        & yes       & yes                  \\
london train           & 24 & 1920x1080  & 385    & 34     & 3207   & 8.3     & static & low       & cloudy  & yes        & no        & yes                  \\
mexico night           & 30 & 640x480    & 1065   & 32     & 8164   & 7.7     & static & low       & night   & yes        & yes       & yes                  \\
milan street           & 24 & 1280x720   & 293    & 40     & 4552   & 15.5    & static & low       & cloudy  & yes        & yes       & yes                  \\
NYC subway gate        & 10 & 640x480    & 555    & 35     & 4578   & 8.2     & static & low       & cloudy  & yes        & no        & yes                  \\
paris bridge           & 30 & 1920x1080  & 674    & 14     & 4506   & 6.7     & static & medium    & cloudy  & no         & yes       & no                   \\
paris street           & 25 & 1280x720   & 300    & 35     & 3240   & 10.8    & moving & medium    & cloudy  & yes        & no        & yes                  \\
pedestrian street      & 50 & 1280x720   & 1200   & 42     & 19799  & 16.5    & static & low       & sunny   & yes        & no        & yes                  \\
phoenix park           & 24 & 1280x720   & 682    & 19     & 7655   & 11.2    & static & low       & sunny   & yes        & no        & no                   \\
red light cross        & 30 & 640x480    & 300    & 16     & 2467   & 8.2     & static & low       & cloudy  & yes        & yes       & no                   \\
regensburg bridge      & 25 & 1280x720   & 638    & 6      & 2967   & 4.7     & static & low       & sunny   & yes        & no        & no                   \\
san franci walkstreet  & 25 & 1280x720   & 300    & 29     & 3659   & 12.2    & static & low       & cloudy  & yes        & yes       & yes                  \\
shanghai overpass      & 24 & 1280x720   & 280    & 50     & 7940   & 28.4    & static & low       & cloudy  & yes        & yes       & yes                  \\
shop corner            & 30 & 1280x720   & 900    & 43     & 10067  & 11.2    & static & medium    & cloudy  & yes        & yes       & yes                  \\
stockholm1             & 25 & 1280x720   & 700    & 35     & 6077   & 8.7     & moving & low       & cloudy  & yes        & no        & yes                  \\
tagbilaran airport     & 13 & 640x480    & 1026   & 39     & 6264   & 6.1     & moving & high      & sunny   & yes        & yes       & no                   \\
taipei underpass       & 30 & 1280x720   & 300    & 26     & 2954   & 9.8     & static & low       & cloudy  & yes        & no        & yes                  \\
tokyo park             & 25 & 1280x720   & 700    & 24     & 6034   & 8.6     & moving & medium    & sunny   & yes        & no        & yes                  \\
tokyo shinkanzen       & 30 & 640x480    & 700    & 49     & 11297  & 16.1    & moving & low       & cloudy  & yes        & no        & yes                  \\
tokyo street           & 25 & 1280x720   & 1000   & 29     & 8359   & 8.4     & moving & low       & sunny   & yes        & no        & yes                  \\
\hline
\multicolumn{3}{|c|}{Total testing}   & 19802  & 1051   & 193497 &      &        &           &         &            &           &                          \\
\hline
\end{tabular}}
\caption{Overview of the testing sequences included in the new dataset.}
\label{datatb:2}
\end{table*}

For the first four datasets evaluation, the proposed Siamese CNN is first pre-trained on the JELMOLI dataset \cite{Ess2} with the loss function in Equation \eqref{eq:2}. For the MOTChallenge \cite{Leal-Taixe03}, the Siamese CNN is first pre-trained on the training set of \cite{Leal-Taixe03}. For the new dataset, the Siamese CNN is first pre-trained on the 10 training sequences. For the regularization parameters in the loss function \eqref{eq:3}, we set $\lambda_0=0.01$, $\lambda=0.02$ and $\eta=0.02$. The weight parameter of the empirical loss is set to $C=0.001$. The learning rate $\beta$ is fixed as 0.01 for all the sequences. The variance $\Sigma$ in the motion-based tracklet affinity model in Equation \eqref{eq:18} is fixed at $\Sigma=diag[625 \ \ 3600]$. A threshold value $\omega$ between 0.5 and 0.6 in Equation \eqref{eq:20} works well for all the datasets. Moreover, a segment of 50 to 80 frames works well for all the sequences.

For fair comparison, the same input detections and groundtruth annotations are utilized for all the trackers in each sequence. Some of the tracking results are directly taken from the corresponding published papers. For the new dataset, we use DPM detector \cite{Felzenszwalb2} to generate the detections. The DPM detections with a score above the threshold value $-0.3$ serve as inputs for all the evaluated trackers in the new dataset.

\subsection{Evaluation Metrics}

We use the popular evaluation metrics defined in \cite{Li}, as well as the CLEAR MOT metrics \cite{Bernardin}: MOTA ($\uparrow$), MOTP ($\uparrow$), Recall ($\uparrow$), Precision ($\uparrow$), False Alarms per Frame (FAF $\downarrow$), False Positives (FP $\downarrow$), False Negatives (FN $\downarrow$), the number of Ground Truth trajectories (GT), Mostly Tracked (MT $\uparrow$), Partially Tracked (PT), Mostly Lost (ML $\downarrow$), the number of Track Fragments (Frag $\downarrow$) and Identity Switches (IDS $\downarrow$). Here, $\uparrow$ denotes higher scores indicate better performance, and $\downarrow$ denotes lower scores indicate better performance.

\subsection{Influence of the Parameters of Segment}

\begin{figure}[!ht]
\centering
\subfigure{
\label{fig:subfig:a} 
\includegraphics[width=0.475\linewidth]{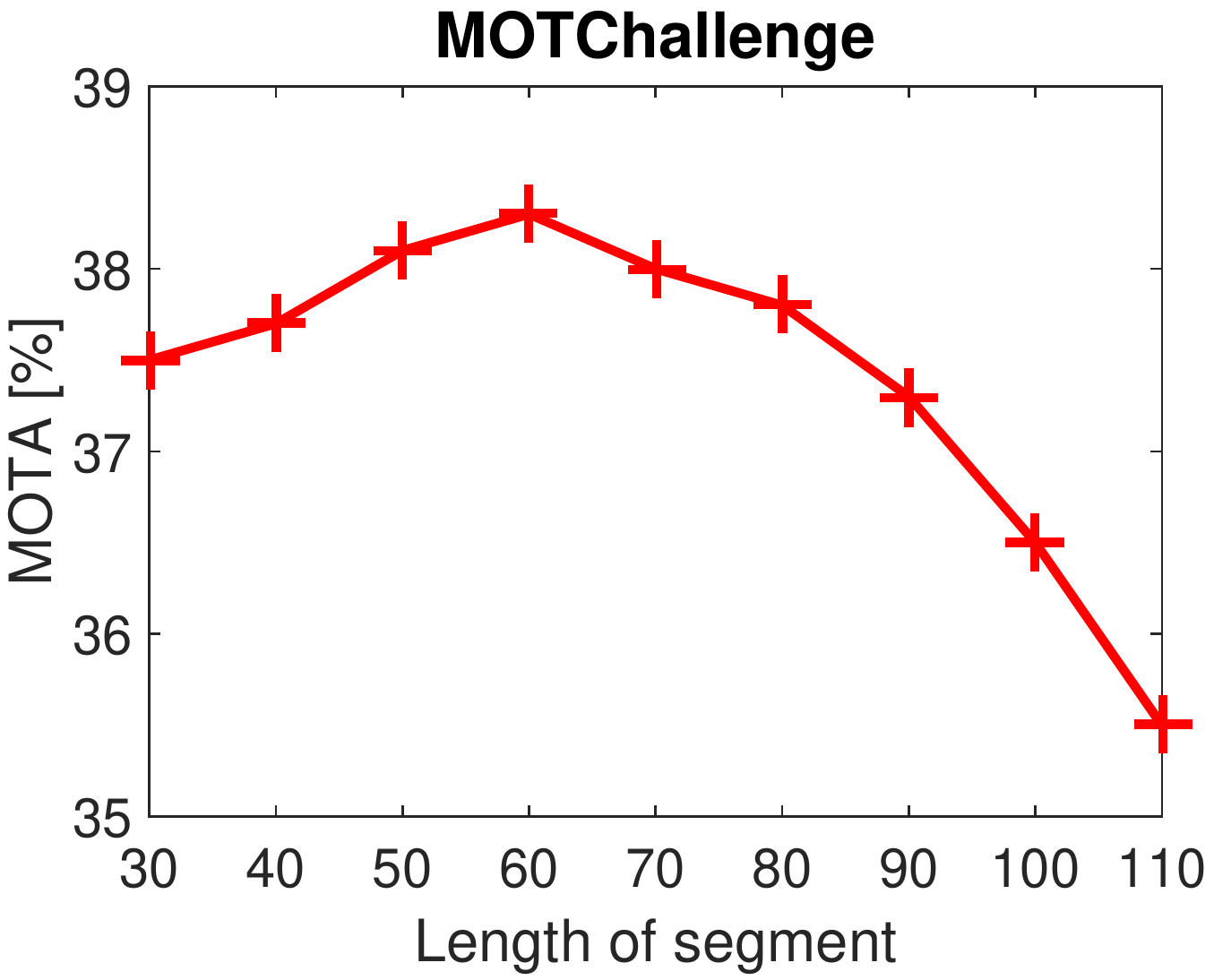}}
\subfigure{
\label{fig:subfig:b} 
\includegraphics[width=0.475\linewidth]{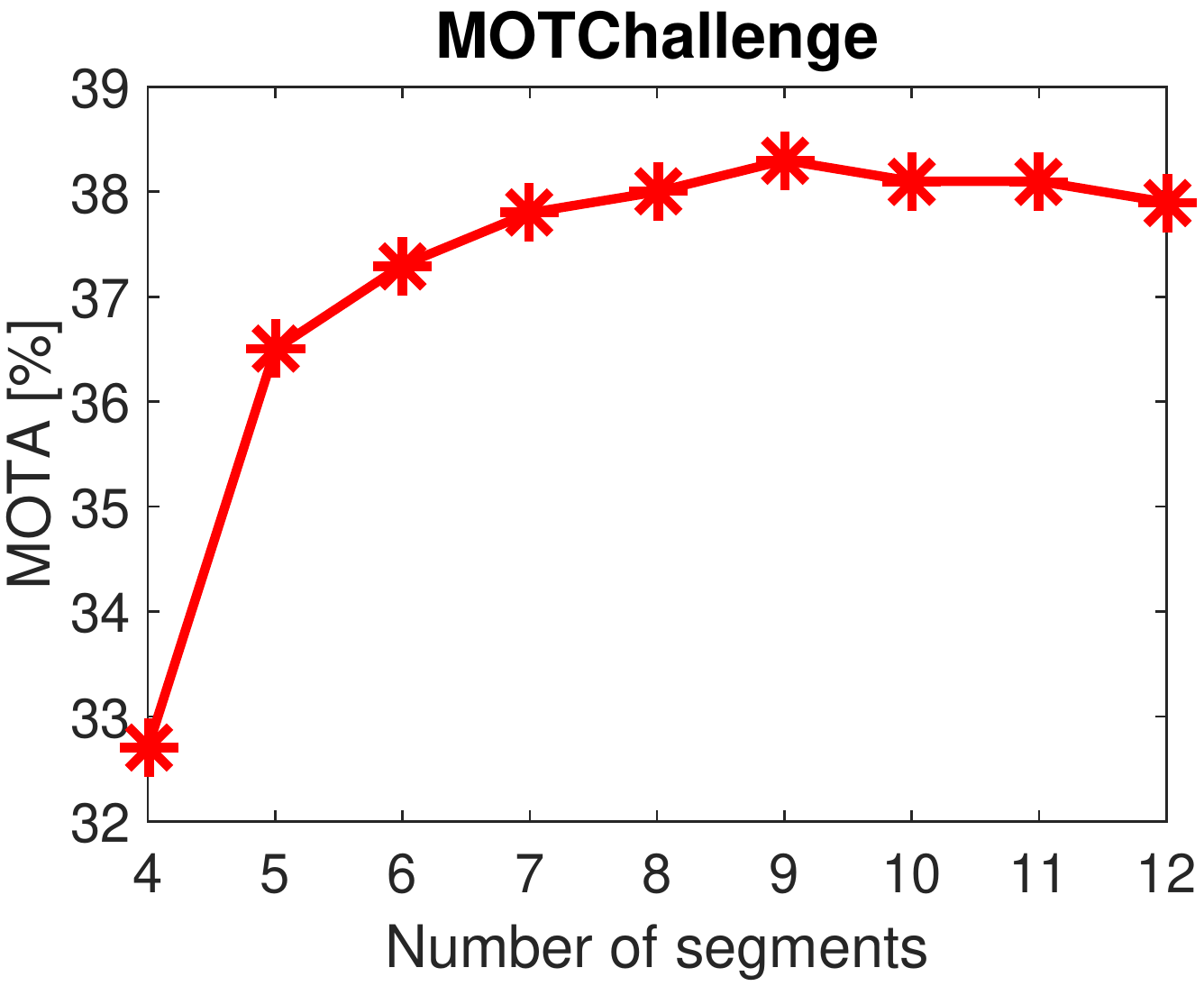}}
\caption{The effects of the segment's parameters on MOTChallenge training set (Measured by MOTA score).}
\label{fig:subfig} 
\end{figure}

The effects of the segment's parameters on the final performance are shown in Figure \ref{fig:subfig}. For the number of segments, it is the average number of all the MOTChallenge training sequences. There is no overlap between segments. From the experiments, it is found that a segment of 50 to 80 frames works well for all the sequences.

\subsection{Performance Evaluation}

\textbf{Evaluation:} To show the effectiveness of joint learning and temporally constrained metrics, two baselines are designed. For \underline{\textbf{Baseline 1}}, the Siamese CNN and the metrics are learned separately. We first learn the Siamese CNN alone by using the loss function \eqref{eq:2}, in which the $M$ is fixed as $M=I$. Then the common metric $M$ is learned separately with the features obtained from the previous learned Siamese CNN. No segment-wise metrics $M_t$ are learned for Baseline 1. For \underline{\textbf{Baseline 2}}, the unified deep model without the temporally constrained multi-task mechanism is learned for tracklet affinity model. In Baseline 2, we use the loss function in Equation \eqref{eq:2} instead of Equation \eqref{eq:3} to learn the unified deep model. Moreover, to show the effectiveness of the CNN fine-tuning and the common metric $M_0$, two more baselines are designed. For \underline{\textbf{Baseline 3}}, the Siamese CNN is pre-trained on JELMOLI dataset but without fine-tuning on target dataset using \eqref{eq:3}. The temporally constrained metrics and the Siamese CNN are learned separately. For \underline{\textbf{Baseline 4}}, no common metric $M_0$ is used. The objective function \eqref{eq:3} without the first term is used for this baseline. Note that the Siamese CNNs of all the baselines are pre-trained on JELMOLI dataset \cite{Ess2}.

From Table \ref{tb:1}, \ref{tb:2}, \ref{tb:3} and \ref{tb:4}, it is found that Baseline 2 achieves overall better performance than Baseline 1 on the evaluated datasets, which proves the effectiveness of the joint learning. Moreover, our method achieves significant improvement in performance on the evaluated datasets, compared with Baseline 2, which validates the superiority of our unified deep model with the temporally constrained multi-task learning mechanism. Our method also achieves overall better performance than Baseline 3 and Baseline 4, which demonstrates the effectiveness of fine-tuning and adding the common metric $M_0$.

\begin{table*}\small
\resizebox{\textwidth}{!}{
\begin{tabular}{|l|c|c|c|c|c|c|c|c|c|c|c|c|c|}
\hline
Method & MOTA & MOTP & Recall & Precision & FAF & FP & FN & GT & MT & PT & ML & Frag & IDS \\
\hline
Milan \emph{et al.} \cite{Milan02} & 90.6\% & 80.2\% & 92.4\% & 98.4\% & 0.07 & 59 & 302 & 23 & 91.3\% & 4.4\% & 4.3\% & 6 & 11 \\
Berclaz \emph{et al.} \cite{Berclaz} & 80.3\% & 72.0\% & 83.8\% & 96.3\% & 0.16 & 126 & 641 & 23 & 73.9\% & 17.4\% & 8.7\% & 22 & 13 \\
Andriyenko \emph{et al.} \cite{Andriyenko2} & 86.3\% & 78.7\% & 89.5\% & 97.6\% & 0.11 & 88 & 417 & 23 & 78.3\% & 17.4\% & 4.3\% & 21 & 38 \\
Andriyenko \emph{et al.} \cite{Andriyenko} & 88.3\% & 79.6\% & 90.0\% & 98.7\% & 0.06 & 47 & 396 & 23 & 82.6\% & 17.4\% & 0.0\% & 14 & 18 \\
Pirsiavash \emph{et al.} \cite{Pirsiavash} & 77.4\% & 74.3\% & 81.2\% & 97.2\% & 0.12 & 93 & 742 & 23 & 60.9\% & 34.8\% & 4.3\% & 62 & 57 \\
Wen \emph{et al.} \cite{Wen} & 92.7\% & 72.9\% & 94.4\% & 98.4\% & 0.08 & 62 & 222 & 23 & 95.7\% & 0.0\% & 4.3\% & 10 & 5 \\
Chari \emph{et al.} \cite{Chari} & 85.5\% & 76.2\% & 92.4\% & 94.3\% & - & 262 & 354 & 19 & 94.7\% & 5.3\% & 0.0\% & 74 & 56 \\
\hline
Baseline1 & 93.6\% & 86.3\% & 96.3\% & 97.7\% & 0.13 & 106 & 170 & 19 & 94.7\% & 5.3\% & 0.0\% & 18 & 18 \\
Baseline2 & 94.3\% & 86.4\% & 96.6\% & 97.9\% & 0.12 & 94 & 157 & 19 & 94.7\% & 5.3\% & 0.0\% & 16 & 11 \\
Baseline3 & 94.0\% & 86.3\% & 96.5\% & 97.8\% & 0.13 & 100 & 163 & 19 & 94.7\% & 5.3\% & 0.0\% & 20 & 12 \\
Baseline4 & 94.7\% & 86.4\% & 97.1\% & 97.8\% & 0.13 & 103 & 134 & 19 & 94.7\% & 5.3\% & 0.0\% & 13 & 8 \\
Ours & 95.8\% & 86.4\% & 97.5\% & 98.4\% & 0.09 & 74 & 115 & 19 & 94.7\% & 5.3\% & 0.0\% & 8 & 4 \\
\hline
\end{tabular}}
\caption{Comparison of tracking results between state-of-the-art methods and ours on PETS 2009 dataset.}
\label{tb:1}
\end{table*}

\begin{table*}\small
\resizebox{\textwidth}{!}{
\begin{tabular}{|l|c|c|c|c|c|c|c|c|c|c|c|c|c|}
\hline
Method & MOTA & MOTP & Recall & Precision & FAF & FP & FN & GT & MT & PT & ML & Frag & IDS \\
\hline
Leal-Taixe \emph{et al.} \cite{Leal-Taixe} & 71.3\% & 71.8\% & - & - & - & - & - & 231 & 58.6\% & 34.4\% & 7.0\% & 363 & 165 \\
Zhang \emph{et al.} \cite{Zhang} &  69.1\% & 72.0\% & - & - & - & - & - & 231 & 53.0\% & 37.7 & 9.3\% & 440 & 243 \\
Benfold \emph{et al.} \cite{Benfold} & 64.3\% & 80.2\% & - & - & - & - & - & 231 & 67.4\% & 26.1\% & 6.5\% & 343 & 222 \\
Pellegrini \emph{et al.} \cite{Pellegrini} & 65.5\% & 71.8\% & - & - & - & - & - & 231 & 59.1\% & 33.9\% & 7.0\% & 499 & 288 \\
Wu \emph{et al.} \cite{Wu01} & 69.5\% & 68.7\% & - & - & - & - & - & 231 & 64.7\% & 27.4\% & 7.9\% & 453 & 209 \\
Yamaguchi \emph{et al.} \cite{Yamaguchi} & 66.6\% & 71.7\% & - & - & - & - & - & 231 & 58.1\% & 35.4\% & 6.5\% & 492 & 302 \\
Possegger \emph{et al.} \cite{Possegger} & 70.7\% & 68.6\% & - & - & - & - & - & 231 & 56.3\% & 36.3\% & 7.4\% & 321 & 157 \\
\hline
Baseline1 & 54.8\% & 72.5\% & 71.1\% &  85.0\% & 1.99 & 895 & 2068 & 231 & 58.0\% & 31.2\% & 10.8\% & 360 & 268\\
Baseline2 & 58.4\% & 73.0\% & 72.2\% &  87.5\% & 1.63 & 735 & 1983 & 231 & 59.7\% & 30.3\% & 10.0\% & 325 & 251\\
Baseline3 & 57.1\% & 72.8\% & 72.3\% &  86.5\% & 1.79 & 806 & 1979 & 231 & 58.9\% & 30.7\% & 10.4\% & 326 & 265\\
Baseline4 & 63.8\% & 74.0\% & 73.2\% &  90.8\% & 1.18 & 530 & 1915 & 231 & 62.8\% & 29\% & 8.2\% & 223 & 153\\
Ours & 67.2\% & 74.5\% & 75.2\% & 92.6\% & 0.95 & 428 & 1770 & 231 & 65.8\% & 27.7\% & 6.5\% & 173 & 146 \\
\hline
\end{tabular}}
\caption{Comparison of tracking results between state-of-the-art methods and ours on Town Centre dataset.}
\label{tb:2}
\end{table*}

\begin{table*}\small
\resizebox{\textwidth}{!}{
\begin{tabular}{|l|c|c|c|c|c|c|c|c|c|c|c|c|c|}
\hline
Method & MOTA & MOTP & Recall & Precision & FAF & FP & FN & GT & MT & PT & ML & Frag & IDS \\
\hline
Shu \emph{et al.} \cite{Shu} & 74.1\% & 79.3\% & 81.7\% & 91.3\% & - & - & - & 14 & - & - & - & - & - \\
Andriyenko \emph{et al.} \cite{Andriyenko2} & 60.0\% & 70.7\% & 69.3\% & 91.3\% & 0.65 & 162 & 756 & 14 & 21.4\% & 71.5\% & 7.1\% & 97 & 68 \\
Andriyenko \emph{et al.} \cite{Andriyenko} & 73.1\% & 76.5\% & 86.8\% & 89.4\% & 1.01 & 253 & 326 & 14 & 78.6\% & 21.4\% & 0.0\% & 70 & 83 \\
Pirsiavash \emph{et al.} \cite{Pirsiavash} & 65.7\% & 75.3\% & 69.4\% & 97.8\% & 0.16 & 39 & 754 & 14 & 7.1\% & 85.8\% & 7.1\% & 60 & 52 \\
Wen \emph{et al.} \cite{Wen} & 88.4\% & 81.9\% & 90.8\% & 98.3\% & 0.16 & 39 & 227 & 14 & 78.6\% & 21.4\% & 0.0\% & 23 & 21 \\
\hline
Baseline1 & 76.5\% & 72.8\% & 86.0\% &  91.6\% & 0.78 & 195 & 344 & 14 & 71.4\% & 28.6\% & 0.0\% & 95 & 39\\
Baseline2 & 80.7\% & 72.6\% & 89.5\% &  92.3\% & 0.74 & 185 & 258 & 14 & 78.6\% & 21.4\% & 0.0\% & 63 & 33\\
Baseline3 & 79.7\% & 72.7\% & 89.1\% &  91.8\% & 0.78 & 196 & 269 & 14 & 78.6\% & 21.4\% & 0.0\% & 70 & 34\\
Baseline4 & 81.9\% & 72.7\% & 89.7\% &  93.1\% & 0.65 & 163 & 254 & 14 & 78.6\% & 21.4\% & 0.0\% & 59 & 30\\
Ours & 85.7\% & 72.9\% & 92.4\% & 93.5\% & 0.63 & 158 & 188 & 14 & 78.6\% & 21.4\% & 0.0\% & 49 & 6 \\
\hline
\end{tabular}}
\caption{Comparison of tracking results between state-of-the-art methods and ours on ParkingLot dataset.}
\label{tb:3}
\end{table*}

\begin{table*}\small
\resizebox{\textwidth}{!}{
\begin{tabular}{|l|c|c|c|c|c|c|c|c|c|c|c|c|c|}
\hline
Method & MOTA & MOTP & Recall & Precision & FAF & FP & FN & GT & MT & PT & ML & Frag & IDS \\
\hline
Kuo \emph{et al.} \cite{Kuo} & - & - & 76.8\% & 86.6\% & 0.891 & - & - & 125 & 58.4\% & 33.6\% & 8.0\% & 23 & 11 \\
Yang \emph{et al.} \cite{Yang2} & - & - & 79.0\% & 90.4\% & 0.637 & - & - & 125 & 68.0\% & 24.8\% & 7.2\% & 19 & 11 \\
Milan \emph{et al.} \cite{Milan} & - & - & 77.3\% & 87.2\% & - & - & - & 125 & 66.4\% & 25.4\% & 8.2\% & 69 & 57 \\
Leal-Taixe \emph{et al.} \cite{Leal-Taixe02} & - & - & 83.8\% & 79.7\% & - & - & - & 125 & 72.0\% & 23.3\% & 4.7\% & 85 & 71 \\
Bae \emph{et al.} \cite{Bae} & 72.03\% & 64.01\% & - & - & - & - & - & 126 & 73.8\% & 23.8\% & 2.4\% & 38 & 18 \\
\hline
Baseline1 & 68.6\% & 76.8\% & 76.7\% & 90.7\% & 0.60 & 812 & 2406 & 125 & 56.8\% & 32.8\% & 10.4\% & 133 & 31 \\
Baseline2 & 71.2\% & 77.0\% & 78.4\% & 91.8\% & 0.54 & 728 & 2236 & 125 & 60.8\% & 29.6\% & 9.6\% & 75 & 20 \\
Baseline3 & 69.6\% & 76.9\% & 77.2\% & 91.3\% & 0.56 & 764 & 2358 & 125 & 58.4\% & 31.2\% & 10.4\% & 115 & 28 \\
Baseline4 & 72.6\% & 77.2\% & 79.1\% & 92.6\% & 0.48 & 653 & 2161 & 125 & 62.4\% & 28.8\% & 8.8\% & 52 & 18 \\
Ours & 75.4\% & 77.5\% & 80.2\% & 94.5\% & 0.36 & 486 & 2050 & 125 & 68.8\% & 24.8\% & 6.4\% & 36 & 6 \\
\hline
\end{tabular}}
\caption{Comparison of tracking results between state-of-the-art methods and ours on ETH dataset.}
\label{tb:4}
\end{table*}

\begin{table*}[!ht]\tiny
\resizebox{\textwidth}{!}{
\begin{tabular}{|l|c|c|c|c|c|c|c|c|c|c|c|}
\hline
Method & MOTA & MOTP & FAF & FP & FN & GT & MT & PT & ML & Frag & IDS \\
\hline
DP\_NMS \cite{Pirsiavash} & 14.5\% & 70.8\% & 2.3 & 13,171 & 34,814 & 721 & 6.0\% & 53.2\% & 40.8\% & 3090 & 4537 \\
SMOT \cite{Dicle} & 18.2\% & 71.2\% & 1.5 & 8,780 & 40,310 & 721 & 2.8\% & 42.4\% & 54.8\% & 2132 & 1148 \\
CEM \cite{Milan02} & 19.3\% & 70.7\% & 2.5 & 14,180 & 34,591 & 721 & 8.5\% & 45\% & 46.5\% & 1023 & 813 \\
TC\_ODAL \cite{Bae} & 15.1\% & 70.5\% & 2.2 & 12,970 & 38,538 & 721 & 3.2\% & 41\% & 55.8\% & 1716 & 637 \\
ELP \cite{McLaughlin} & 25.0\% & 71.2\% & 1.3 & 7,345 & 37,344 & 721 & 7.5\% & 48.7\% & 43.8\% & 1804 & 1396 \\
\hline
CNNTCM (Ours) & 29.6\% & 71.8\% & 1.3 & 7,786 & 34,733 &721 & 11.2\% & 44.8\% & 44.0\% & 943 & 712 \\
\hline
\end{tabular}}
\caption{Comparison of tracking results between state-of-the-art methods and ours on MOTChallenge Benchmark.}
\label{tb:5}
\end{table*}

\begin{table*}[!ht]\small
\resizebox{\textwidth}{!}{
\begin{tabular}{|l|c|c|c|c|c|c|c|c|c|c|c|c|c|c|}
\hline
Method & MOTA & MOTP & Recall & Precision & FAF & FP & FN & GT & MT & PT & ML & Frag & IDS \\
\hline
DP\_NMS \cite{Pirsiavash} & 29.7\% & 75.2\% & 31.4\% & 97.2\% & 0.09 & 1,742 & 130,338 & 1051 & 6.7\% & 37.4\% & 55.9\% & 1,738 & 1,359 \\
SMOT \cite{Dicle} & 33.8\% & 73.9\% & 43.5\% & 85.3\% & 0.72 & 14,248 & 107,310 & 1051 & 6.2\% & 58.1\% & 35.7\% & 5,507 & 4,214 \\
CEM \cite{Milan02} & 32.0\% & 74.2\% & 39.7\% & 85.1\% & 0.67 & 13,185 & 114,576 & 1051 & 11.1\% & 42.4\% & 46.5\% & 2,016 & 1,367 \\
TC\_ODAL \cite{Bae} & 32.8\% & 73.2\% & 56.2\% & 71.3\% & 2.17 & 42,913 & 83,206 & 1051 & 21.9\% & 48.1\% &30.0\% & 3,655 & 1,577 \\
ELP \cite{McLaughlin} & 34.1\% & 75.9\% & 40.2\% & 89.2\% & 0.47 & 9,237 & 113,590 & 1051 & 9.4\% & 43.9\% & 46.7\% & 2,451 & 2,301 \\
\hline
CNNTCM (Ours) & 39.0\% & 74.5\% & 41.7\% & 95.1\% & 0.20 & 4,058 & 110,635 & 1051 & 11.9\% & 43.2\% & 44.9\% & 1,946 & 1,236 \\
\hline
\end{tabular}}
\caption{Comparison of tracking results between state-of-the-art methods and ours on the new dataset.}
\label{tb:6}
\end{table*}

\begin{figure}[t]
\centering
\subfigure[Camera motion]{
\label{fig:subfig_06:a} 
\includegraphics[width=0.475\linewidth]{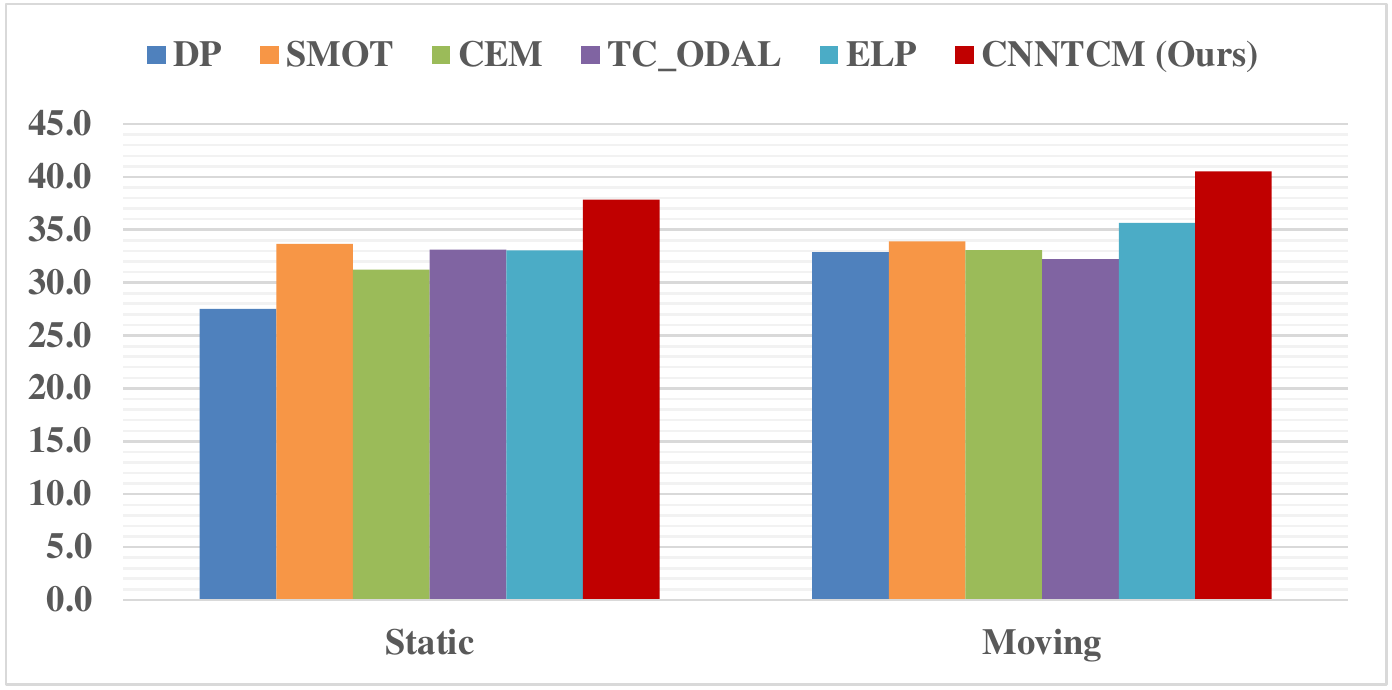}}
\subfigure[Viewpoint]{
\label{fig:subfig_06:b} 
\includegraphics[width=0.475\linewidth]{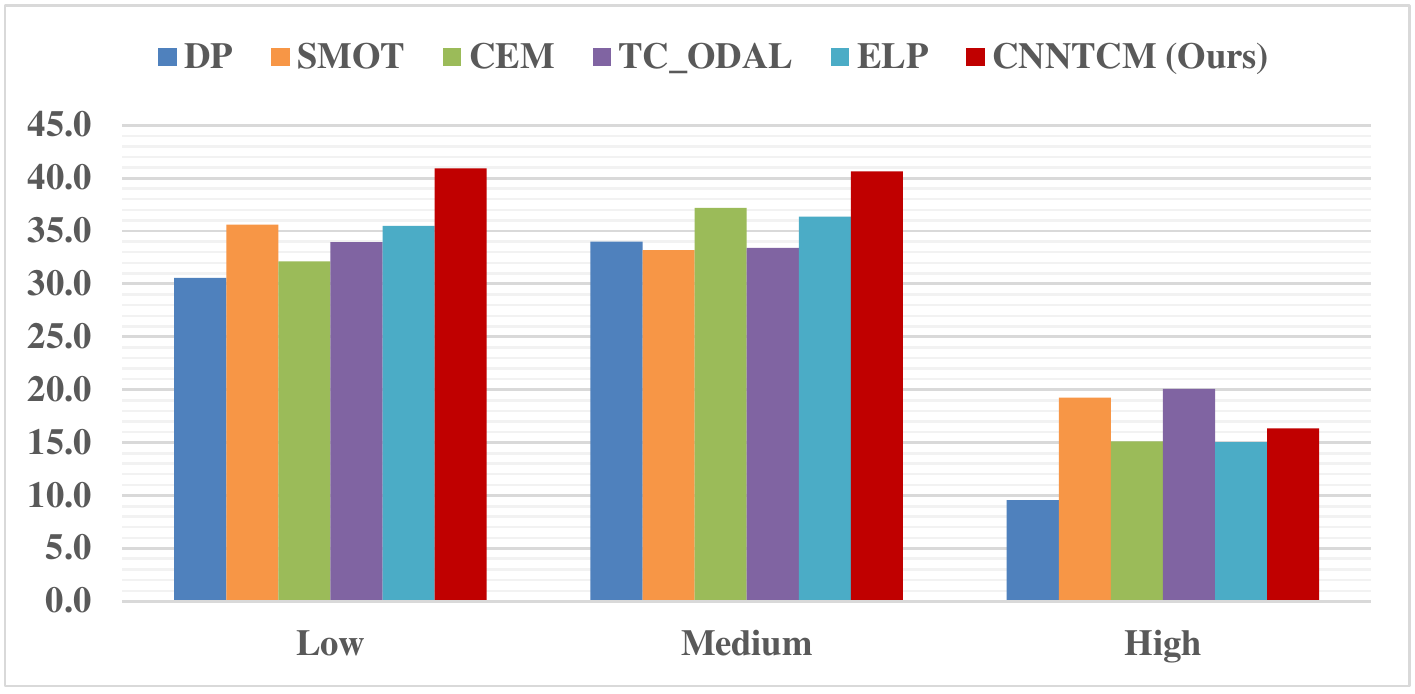}}
\subfigure[Weather conditions]{
\label{fig:subfig_06:c} 
\includegraphics[width=0.475\linewidth]{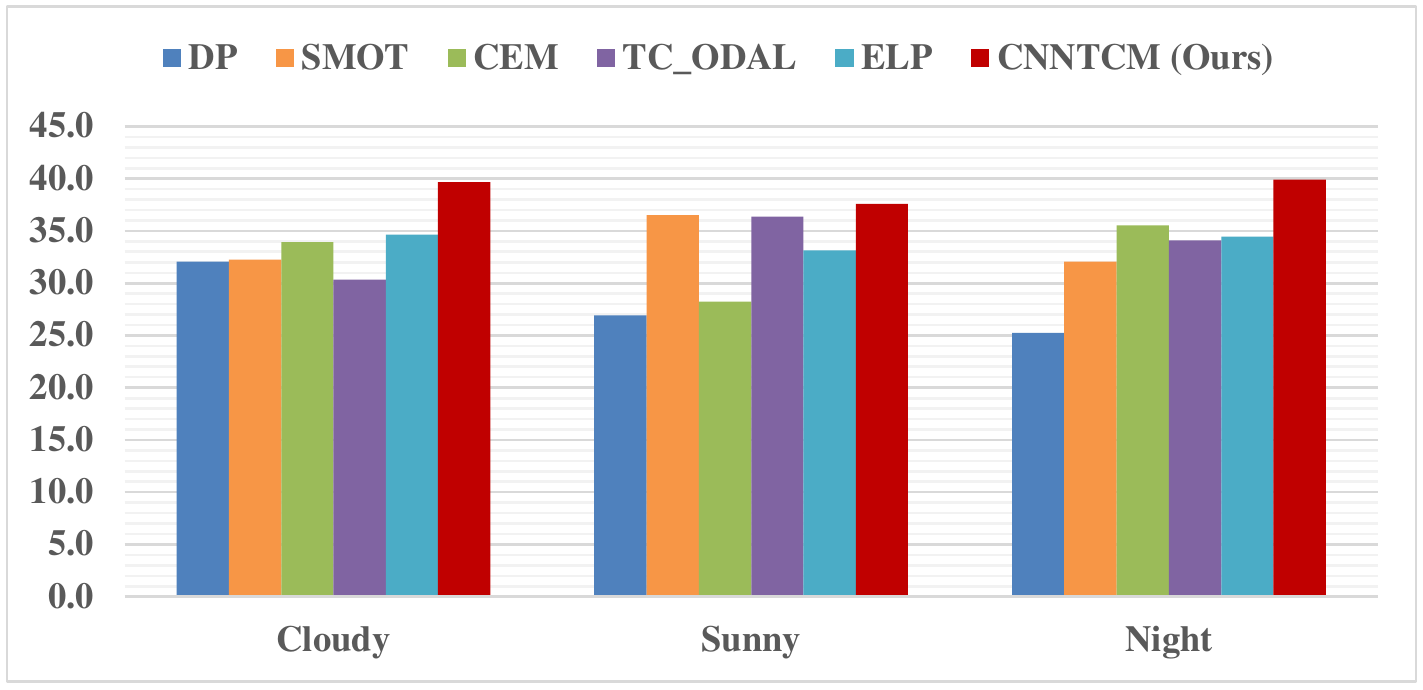}}
\subfigure[Pedestrian view]{
\label{fig:subfig_06:d} 
\includegraphics[width=0.475\linewidth]{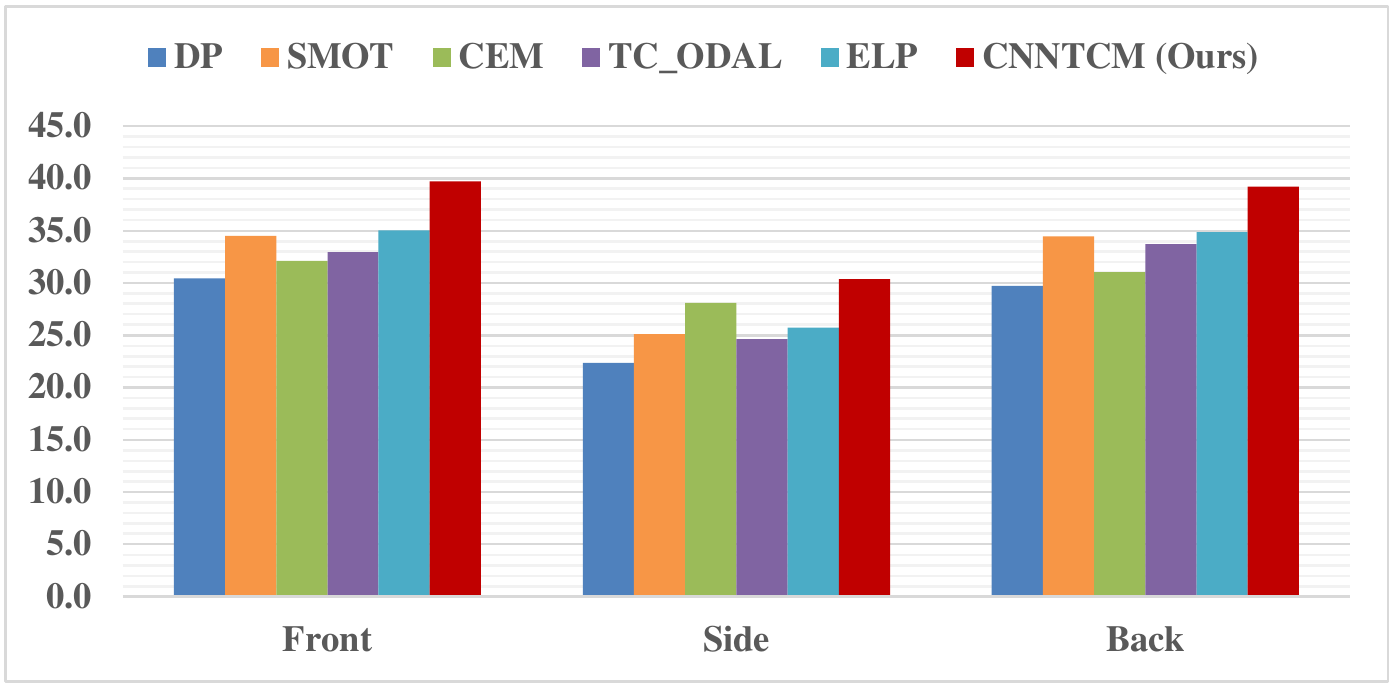}}
\caption{Comparison of the evaluated tracking methods under different conditions in terms of MOTA score (\%).}
\label{fig:subfig_06} 
\end{figure}

We further evaluate our method on the recent MOTChallenge 2D Benchmark \cite{Leal-Taixe03}. The qualitative results of our method (CNNTCM) are available on the MOTChallenge Benchmark website \cite{MOTChallenge}. From Table \ref{tb:5}, it is found that our method achieves better performance on all evaluation measures compared with a recent work \cite{Dicle} which is also based on the GLA framework. Compared with other state-of-the-art methods, our method achieves better or comparable performance on all the evaluation measures.

Moreover, to further show the generality and effectiveness of the proposed method on large-scale sequences, a new dataset with 40 diverse sequences is built for performance evaluation. 5 state-of-the-art tracking methods with released source codes are used in evaluation for the new dataset. The parameters of each evaluated tracking method are fine-tuned on the 10 training sequences. As shown in Table \ref{tb:6}, our method achieves the best performance on MOTA and IDS, which are the most two direct measures for tracklet association evaluation, among all the evaluated tracking methods. The comparison of tracking results between these five state-of-the-art methods and ours on the new dataset under different conditions is shown in Figure \ref{fig:subfig_06}.

\textbf{Computational speed:} Our system was implemented using the MatConvNet toolbox \cite{Vedaldi} on a server with a 2.60GHz CPU and a Tesla K20c GPU. The computation speed is subject to the number of targets in a video sequence. The speeds of our method are about 0.38, 0.81, 0.50, 0.60, 0.59, 0.55 (sec/frame) for PETS 2009, Town Centre, ParkingLot, ETH, MOTChallenge, and the new dataset, respectively, excluding the detection step. Note that speed-up can be achieved by further optimization of the codes.

\section{Conclusion} \label{sec:5}

In this paper, a novel unified deep model for tracklet association is presented. This deep model can jointly learn the Siamese CNN and temporally constrained metrics for tracklet affinity models. The experimental results of Baseline 1 and Baseline 2 validate the effectiveness of the joint learning and the temporally constrained multi-task learning mechanism of the proposed unified deep model. Baseline 3 and Baseline 4 demonstrate the effectiveness of fine-tuning and adding the common metric. Moreover, a new large-scale dataset with 40 fully annotated sequences is created to facilitate multi-target tracking evaluation. Furthermore, extensive experimental results on five public datasets and the new large-scale dataset compared with state-of-the-art methods also demonstrate the superiority of our method.

\appendices

\section*{Acknowledgment}

The authors would like to acknowledge the Research Scholarship from School of Electrical and Electronics Engineering, Nanyang Technological University, Singapore.

\ifCLASSOPTIONcaptionsoff
  \newpage
\fi



%


{\small
\bibliographystyle{IEEEtran}
\bibliography{egbib}

\begin{thebibliography}{10}
\providecommand{\url}[1]{#1}
\csname url@samestyle\endcsname
\providecommand{\newblock}{\relax}
\providecommand{\bibinfo}[2]{#2}
\providecommand{\BIBentrySTDinterwordspacing}{\spaceskip=0pt\relax}
\providecommand{\BIBentryALTinterwordstretchfactor}{4}
\providecommand{\BIBentryALTinterwordspacing}{\spaceskip=\fontdimen2\font plus
\BIBentryALTinterwordstretchfactor\fontdimen3\font minus
  \fontdimen4\font\relax}
\providecommand{\BIBforeignlanguage}[2]{{%
\expandafter\ifx\csname l@#1\endcsname\relax
\typeout{** WARNING: IEEEtran.bst: No hyphenation pattern has been}%
\typeout{** loaded for the language `#1'. Using the pattern for}%
\typeout{** the default language instead.}%
\else
\language=\csname l@#1\endcsname
\fi
#2}}
\providecommand{\BIBdecl}{\relax}
\BIBdecl

\bibitem{Dalal}
N.~Dalal and B.~Triggs, ``Histograms of oriented gradients for human
  detection,'' in \emph{\emph{CVPR}}, 2005.

\bibitem{Wang1}
X.~Wang, T.~X. Han, and S.~Yan, ``An hog-lbp human detector with partial
  occlusion handling,'' in \emph{\emph{ICCV}}, 2009.

\bibitem{Felzenszwalb2}
P.~Felzenszwalb, R.~Girshick, D.~McAllester, and D.~Ramanan, ``Object detection
  with discriminatively trained part based models,'' \emph{\emph{IEEE
  Transactions on Pattern Analysis and Machine Intelligence}}, vol.~32, no.~9,
  pp. 1627--1645, 2010.

\bibitem{Kuo}
C.~H. Kuo and R.~Nevatia, ``How does person identity recognition help
  multi-person tracking?'' in \emph{\emph{CVPR}}, 2011.

\bibitem{Yang2}
B.~Yang and R.~Nevatia, ``An online learned crf model for multi-target
  tracking,'' in \emph{\emph{CVPR}}, 2012.

\bibitem{Dicle}
C.~Dicle, O.~Camps, and M.~Sznaier, ``The way they move: tracking multiple
  targets with similar appearance,'' in \emph{\emph{ICCV}}, 2013.

\bibitem{Wen}
L.~Wen, W.~Li, J.~Yan, Z.~Lei, D.~Yi, and S.~Z. Li, ``Multiple target tracking
  based on undirected hierarchical relation hypergraph,'' in
  \emph{\emph{CVPR}}, 2014.

\bibitem{Wang111}
B.~Wang, G.~Wang, K.~L. Chan, and L.~Wang, ``Tracklet association with online
  target-specific metric learning,'' in \emph{\emph{CVPR}}, 2014.

\bibitem{Bae}
S.~H. Bae and K.~J. Yoon, ``Robust online multi-object tracking based on
  tracklet confidence and online discriminative appearance learning,'' in
  \emph{\emph{CVPR}}, 2014.

\bibitem{Krizhevsky}
A.~Krizhevsky, I.~Sutskever, and G.~Hinton, ``Imagenet classification with deep
  convolutional neural networks,'' in \emph{\emph{NIPS}}, 2012.

\bibitem{Girshick}
R.~Girshick, J.~Donahue, T.~Darrell, and J.~Malik, ``Rich feature hierarchies
  for accurate object detection and semantic segmentation,'' in
  \emph{\emph{CVPR}}, 2014.

\bibitem{Shmoys}
D.~Shmoys and E.~Tardos, ``An approximation algorithm for the generalized
  assignment problem,'' \emph{\emph{Mathematical Programming}}, vol.~62, no.~1,
  pp. 461--474, 1993.

\bibitem{Gold}
S.~Gold and A.~Rangarajan, ``Softmax to softassign: Neural network algorithms
  for combinatorial optimization,'' \emph{\emph{Journal of Artificial Neural
  Nets}}, vol.~2, no.~4, pp. 381--399, 1995.

\bibitem{Oquab}
M.~Oquab, L.~Bottou, I.~Laptev, and J.~Sivic, ``Learning and transferring
  mid-level image representations using convolutional neural networks,'' in
  \emph{\emph{CVPR}}, 2014.

\bibitem{Wang2}
N.~Wang and D.~Y. Yeung, ``Learning a deep compact image representation for
  visual tracking,'' in \emph{\emph{NIPS}}, 2013.

\bibitem{Li01}
H.~Li, Y.~Li, and F.~Porikli, ``Deeptrack: Learning discriminative feature
  representations by convolutional neural networks for visual tracking,'' in
  \emph{\emph{BMVC}}, 2014.

\bibitem{Wang3}
L.~Wang, T.~Liu, G.~Wang, K.~L. Chan, and Q.~Yang, ``Video tracking using
  learned hierarchical features,'' \emph{\emph{IEEE Transactions on Image
  Processing}}, vol.~24, no.~4, pp. 1424--1435, 2015.

\bibitem{Zhang03}
K.~Zhang, Q.~Liu, Y.~Wu, and M.~H. Yang, ``Robust tracking via convolutional
  networks without learning,'' \emph{\emph{arXiv preprint}}, vol.
  arXiv:1501.04505, 2015.

\bibitem{Hong01}
S.~Hong, T.~You, S.~Kwak, and B.~Han, ``Online tracking by learning
  discriminative saliency map with convolutional neural network,''
  \emph{\emph{arXiv preprint}}, vol. arXiv:1502.06796, 2015.

\bibitem{Huang2}
C.~Huang, B.~Wu, and R.~Nevatia, ``Robust object tracking by hierarchical
  association of detection responses,'' in \emph{\emph{ECCV}}, 2008.

\bibitem{LeCun}
Y.~LeCun, L.~Bottou, Y.~Bengio, and P.~Haffner, ``Gradient-based learning
  applied to document recognition,'' \emph{\emph{Proc. of the IEEE}}, 1998.

\bibitem{Zhang}
L.~Zhang, Y.~Li, and R.~Nevatia, ``Global data association for multi-object
  tracking using network flows,'' in \emph{\emph{CVPR}}, 2008.

\bibitem{Pirsiavash}
H.~Pirsiavash, D.~Ramanan, and C.~Fowlkes, ``Globally-optimal greedy algorithms
  for tracking a variable number of objects,'' in \emph{\emph{CVPR}}, 2011.

\bibitem{Butt}
A.~Butt and R.~T. Collins, ``Multi-target tracking by lagrangian relaxation to
  min-cost network flow,'' in \emph{\emph{CVPR}}, 2013.

\bibitem{Ferryman}
J.~Ferryman and A.~Shahrokni, ``Pets2009: Dataset and challenge,'' in
  \emph{\emph{Winter-PETS}}, 2009.

\bibitem{Benfold}
B.~Benfold and I.~Reid, ``Stable multi-target tracking in real-time
  surveillance video,'' in \emph{\emph{CVPR}}, 2011.

\bibitem{Shu}
G.~Shu, A.~Dehghan, O.~Oreifej, E.~Hand, and M.~Shah, ``Part-based
  multiple-person tracking with partial occlusion handling,'' in
  \emph{\emph{CVPR}}, 2012.

\bibitem{Ess}
A.~Ess, K.~S.~B. Leibe, and L.~van Gool, ``Robust multiperson tracking from a
  mobile platform,'' \emph{\emph{IEEE Transactions on Pattern Analysis and
  Machine Intelligence}}, vol.~31, no.~10, pp. 1831--1846, 2009.

\bibitem{Leal-Taixe03}
L.~Leal-Taixe, A.~Milan, I.~Reid, S.~Roth, and K.~Schindler, ``Motchallenge
  2015: Towards a benchmark for multi-target tracking,'' \emph{\emph{arXiv
  preprint}}, vol. arXiv: 1504.01942, 2015.

\bibitem{Andriluka2}
M.~Andriluka, S.~Roth, and B.~Schiele, ``People-tracking-by-detection and
  people-detection-by-tracking,'' in \emph{\emph{CVPR}}, 2008.

\bibitem{Geiger01}
A.~Geiger, P.~Lenz, and R.~Urtasun, ``Are we ready for autonomous driving? the
  kitti vision benchmark suite,'' in \emph{\emph{CVPR}}, 2012.

\bibitem{Wuyi01}
Y.~Wu, J.~Lim, and M.~H. Yang, ``Online object tracking: a benchmark,'' in
  \emph{\emph{CVPR}}, 2013.

\bibitem{Ess2}
A.~Ess, B.~Leibe, and L.~van Gool, ``Depth and appearance for mobile scene
  analysis,'' in \emph{\emph{ICCV}}, 2007.

\bibitem{Li}
Y.~Li, C.~Huang, and R.~Nevatia, ``Learning to associate: Hybridboosted
  multi-target tracker for crowded scene,'' in \emph{\emph{CVPR}}, 2009.

\bibitem{Bernardin}
K.~Bernardin and R.~Stiefelhagen, ``Evaluating multiple object tracking
  performance: The clear mot metrics,'' \emph{\emph{EURASIP J. Image and Video
  Processing}}, 2008.

\bibitem{Milan02}
A.~Milan, S.~Roth, and K.~Schindler, ``Continuous energy minimization for
  multi-target tracking,'' \emph{\emph{IEEE Transactions on Pattern Analysis
  and Machine Intelligence}}, vol.~36, no.~1, pp. 58--72, 2014.

\bibitem{Berclaz}
J.~Berclaz, F.~Fleuret, E.~Turetken, and P.~Fua, ``Multiple object tracking
  using k-shortest paths optimization,'' \emph{\emph{IEEE Transactions on
  Pattern Analysis and Machine Intelligence}}, vol.~33, no.~9, pp. 1806--1819,
  2011.

\bibitem{Andriyenko2}
A.~Andriyenko and K.~Schindler, ``Multi-target tracking by continuous energy
  minimization,'' in \emph{\emph{CVPR}}, 2011.

\bibitem{Andriyenko}
A.~Andriyenko, K.~Schindler, and S.~Roth, ``Discrete-continuous optimization
  for multi-target tracking,'' in \emph{\emph{CVPR}}, 2012.

\bibitem{Chari}
V.~Chari, S.~L. Julien, I.~Laptev, and J.~Sivic, ``On pairwise costs for
  network flow multi-object tracking,'' in \emph{\emph{CVPR}}, 2015.

\bibitem{Leal-Taixe}
L.~Leal-Taixe, G.~Pons-Moll, and B.~Rosenhahn, ``Everybody needs somebody:
  Modeling social and grouping behavior on a linear programming multiple people
  tracker,'' in \emph{\emph{ICCV Workshops}}, 2011.

\bibitem{Pellegrini}
S.~Pellegrini, A.~Ess, K.~Schindler, and L.~V. Gool, ``You'll never walk alone:
  modeling social behavior for multi-target tracking,'' in \emph{\emph{ICCV}},
  2009.

\bibitem{Wu01}
Z.~Wu, J.~Zhang, and M.~Betke, ``Online motion agreement tracking,'' in
  \emph{\emph{BMVC}}, 2013.

\bibitem{Yamaguchi}
K.~Yamaguchi, A.~C. Berg, L.~E. Ortiz, and T.~L. Berg, ``Who are you with and
  where are you going?'' in \emph{\emph{CVPR}}, 2011.

\bibitem{Possegger}
H.~Possegger, T.~Mauthner, P.~M. Roth, and H.~Bischof, ``Occlusion geodesics
  for online multi-object tracking,'' in \emph{\emph{CVPR}}, 2014.

\bibitem{Milan}
A.~Milan, K.~Schindler, and S.~Roth, ``Detection- and trajectory-level
  exclusion in multiple object tracking,'' in \emph{\emph{CVPR}}, 2013.

\bibitem{Leal-Taixe02}
L.~Leal-Taixe, M.~Fenzi, A.~Kuznetsova, B.~Rosenhahn, and S.~Savarese,
  ``Learning an image-based motion context for multiple people tracking,'' in
  \emph{\emph{CVPR}}, 2014.

\bibitem{McLaughlin}
N.~McLaughlin, J.~M.~D. Rincon, and P.~Miller, ``Enhancing linear programming
  with motion modeling for multi-target tracking,'' in \emph{\emph{WACV}},
  2015.

\bibitem{MOTChallenge}
``Multiple object tracking benchmark,'' \url{http://motchallenge.net/}.

\bibitem{Vedaldi}
A.~Vedaldi and K.~Lenc, ``Matconvnet -- convolutional neural networks for
  matlab,'' \emph{\emph{arXiv preprint}}, vol. arXiv:1412.4564, 2014.

\end{thebibliography}
}

\end{document}